\documentclass[acmtog, authorversion]{acmart}

\AtBeginDocument{%
  }

\setcopyright{cc}
\setcctype{by}
\acmJournal{TOG}
\acmYear{2025} 
\acmVolume{0} 
\acmNumber{0} 
\acmArticle{0} 
\acmMonth{0} 
\acmPrice{}
\acmDOI{10.1145/3736717}

\usepackage{amsmath}
\usepackage{algorithmic}
\usepackage{algorithm}
\usepackage{array}
\usepackage[caption=false,font=normalsize,labelfont=sf,textfont=sf]{subfig}
\usepackage{textcomp}
\usepackage{stfloats}
\usepackage{url}
\usepackage{verbatim}
\usepackage{graphicx}
\usepackage{cite}

\usepackage{mathtools}
\usepackage{adjustbox}
\usepackage{multirow} 
\usepackage{xurl} 

\usepackage{subfig}
\usepackage{pgf,tikz,pgfplots}
\pgfplotsset{compat=1.15}
\usepackage{mathrsfs}
\usetikzlibrary{arrows}
\usepackage{tikz-3dplot} 
\usetikzlibrary{matrix,decorations.pathreplacing} 

\usepackage{booktabs} 

\usepackage{enumitem}

\graphicspath{{./Images/Images/} {Images}  {Images/} {/Images/} {../Images}}

\begin{document}

    \title[Topology Type Estimation of Simulated 4D Image Data by Combining Downscaling and CNNs]{Topology Type Estimation of Simulated 4D Image Data by Combining Downscaling and Convolutional Neural Networks}

    \author{Khalil Mathieu Hannouch}
    \authornote{Both authors contributed equally to this research.}
    \email{khalil.hannouch@uon.edu.au}
    \author{Stephan Chalup}
    \authornotemark[1]
    \email{stephan.chalup@acm.org}
    \orcid{0000-0002-7886-3653}
    \affiliation{%
      \institution{The University of Newcastle}
      \streetaddress{University Drive}
      \city{Callaghan}
      \state{NSW}
      \country{Australia}
      \postcode{2308}
    }
    

    \begin{abstract}
        The topological analysis of four-dimensional (4D) image-type data is challenged by the immense size that these datasets can reach. This can render the direct application of methods, like persistent homology and convolutional neural networks (CNNs), impractical due to computational constraints. 
        This study aims to estimate the topology type of 4D image-type data cubes that exhibit topological intricateness and size above our current processing capacity. The experiments using synthesised 4D data and a real-world 3D data set demonstrate that it is possible to circumvent computational complexity issues by applying downscaling methods to the data before training a CNN. This is achievable even when persistent homology software indicates that downscaling can significantly alter the homology of the training data.
        When provided with downscaled test data, the CNN can still estimate the Betti numbers of the original sample cubes with over 80\% accuracy, which outperforms the persistent homology approach, whose accuracy deteriorates under the same conditions. The accuracy of the CNNs can be further increased by moving from a mathematically-guided approach to a more vision-based approach where cavity types replace the Betti numbers as training targets.
    \end{abstract}



    \received{19 December 2023}

    \maketitle

    \section{Introduction}\label{sec:introduction}
        In scenarios where data inherently exists in four spatial dimensions, conventional methods that reduce this data to 3D or 2D can result in a significant loss of information. This challenge arises, for instance, when residual variance in manifold learning suggests that fully capturing or revealing the essential structure of the data requires at least a 4D representation~\citep{LeeVerleysenBook2007, TenenbaumEtAl2000}. Although there have been few experimental studies exploring cases where understanding the global topology of the involved manifolds necessitates an ambient space of more than three dimensions~\citep{AzizEtAl2019,CarlssonEtAl2008,JoswigEtAl2022}, the importance of 4D topological data analysis (TDA) is underscored by the rich and intricate topological structures found in 4D. These structures, such as those of certain 3-manifolds, can significantly influence the characteristics of dynamical systems residing on them, as articulated by the Poincar\'e-Hopf theorem~\citep{BrasseletEtAl2009}. For the graphics community, 4D TDA represents a promising frontier, offering new avenues for visualizing and manipulating high-dimensional data, and paving the way for innovations in science and engineering that were previously constrained by the limitations of 2D and 3D methodologies.

        An understanding of the topological structure of image-type data can be critical in application areas such as material science~\citep{AlSahlaniEtAl2018, Dua20B} and medicine~\citep{Cang17,Kim19,Lou21}, 
        where methods such as Magnetic Resonance Imaging (MRI) and Computed Tomography (CT) may be used to determine the existence and shape of cavities within materials, or identify normal and pathological anatomical structures.
        By considering an $n$-dimensional image as a manifold with boundary, its structural properties can be investigated from a topological perspective. MRI and CT scans offer examples of images in the 3-dimensional (3D) setting, as does real-time ultrasound (US), which captures 2D images over time to produce 3D image data. 3D images are comprised of voxels, which are the 3D analogue of a pixel.
         
        In the 4D setting, 4D-US, functional-MRI, and 4D-CT offer methods to scan a 3D target over time to produce 4D image data; this affords the observation of processes and movements.
        These data are usually produced by collecting a synchronised sequence of 2D slices, which are then rectified into a 4D format by using slice timing correction techniques that employ various interpolation methods in order to accommodate for the time delay that is exhibited as each slice is captured~\citep{Pau16,Par17}.
        4D data can also occur if 2D or 3D data are equipped with dimensions other than time.
        
        4D imaging in the medical diagnostic arena can allow moving structures to be imaged over time. A review paper by~\citet{Kwo15} offers a broad look into how these dynamic imaging techniques can be used to observe visceral, musculo-skeletal, and vascular structures in order to assess joint instability and valve motion. 
        An analysis of the topological characteristics of medical imaging is also a research consideration~\citep{Lou21}. For example, in cancer research, \citet{Kim19} investigated the impact of TDA in helping to differentiate between MRI scans of subjects with and without a genetic deletion event associated with a better glioma prognosis.

        In material science, micro-tomography has been used to observe the structural evolution of materials while they undergo hydration processes~\citep{Zha22}, and to study the effect of exposure to load, temperature change, or current on manufactured porous materials, such as cellular materials and syntactic foams~\citep{AlSahlaniEtAl2018,Dua20B}.

        Unfortunately, 4D imaging can result in data that are dense, in that they capture vast regions of target material versus empty space, and data that require large amounts of storage. 
        Furthermore, due to the additional dimension of the data in 4D, significant computational-, memory-, and time-complexity challenges of 4D TDA methods must be addressed.

        Currently, the candidate techniques for TDA in the discussed areas are various forms of persistent homology~\citep{Ede10,Ott16}, and more recently also convolutional neural networks (CNNs)~
        \citep{Pau19,Pee23}.

        The present study proposes and demonstrates the feasibility of an approach that combines the downscaling of large 4D image-type manifold data, which comprise of black-and-white toxels (the 4D analogue of a pixel), and the training of a 4D CNN~\citep{Han23}, 
        in order to estimate the topological characteristics of the data. 
        In particular, all four dimensions of the data that we consider are treated equally; one could indeed inspect these data from any 4D perspective, and not necessarily assume that they arise from observing 3D samples as they evolve with time.
        In the context of this work, the term \textit{large} data refers to data that is expensive, or even infeasible, to analyse in its raw form because of computational or memory challenges that are encountered by our currently available hardware (see Sections~\ref{sec:background} and \ref{subsec:dataset_parameters}). 
        
        To corroborate the workings of the approach in 4D, which we currently could only test on synthetic data due to computational constraints, we demonstrate a 3D version of the approach when applied to a real-world 3D scan of a metallic syntactic foam~\citep{Fie20}. While persistent homology can calculate the homology of general data, our approach demonstrates that a CNN combined with downscaling can become a more efficient topology-type estimator for classes of data it has been trained on.
        
        
        While we specifically address the case of image-type data in this work, persistent homology-based methods may be more suitable for data in point-cloud or mesh format, as we discuss in the concluding paragraph of Section~\ref{sec:background}.
        Although data analysis techniques have received significant attention in various areas of science and mathematics, we are still in the early stages of exploring how to apply computer vision and graphics-based approaches to the task of estimating the topological characteristics of data.

        The main contributions that we provide in this work include:
        \begin{enumerate}[label=(\roman*)]
            \item the generation of large synthetic 4D image data samples with non-trivial topologies (Section~\ref{sec:dataset_generation}), 
            
            \item the implementation of a `4D-camera' that was presented in a previous workshop paper~\citep{Han23} and is summarised here in Appendix~\ref{app:4d_camera},
            
            \item training results that demonstrate that 4D CNNs can estimate the topology of the data even after downscaling (Section~\ref{sec:experiments_and_results_4d}), 
            
            \item a comparison of two fundamentally different approaches in 4D, one where CNNs estimate the Betti numbers of the sample directly (Sections~\ref{subsec:downsampling_approach} and \ref{subsec:average-pooling_approach}), like persistent homology, versus one where the CNN classifies the cavities based on their topology type 
            (Section~\ref{subsec:cavity-focused_approach}), 
            
            \item the use of a real-world derived 3D dataset to demonstrate that the `CNN and downscaling'- approach also works on data with real-world features  (Section~\ref{sec:a_study_using_3d})
            
            \item an experimental comparison with a representative persistent homology approach in 4D (Section~\ref{subsec:efficacy}) and 3D (Section~\ref{sec:experiments_and_results_3d}), and
            
            \item a discussion that addresses some current limitations of the proposed approach to topology estimation, including issues that would need to be considered when transferring the approach between different real-world data domains (Section~\ref{sec:discussion}).
        \end{enumerate}
        

        The contributions of the main part of the present paper are completely new but build on a previous workshop paper~\citep{Han23}. All material from the workshop paper that provides necessary or useful context for the present paper has been summarised in Appendices~\ref{app:4d_camera} and~\ref{app_sec:algebraic_topology}.

        
        
    \section{Background}\label{sec:background}
        Inspired by the structure of 3D data blocks in material science~\citep{Fie20}, the present study used simulated 4D data cubes with cavities, which could be described as the 4D analogue of a 3D foam or a block of Swiss cheese. The boundaries of the cavities in such 4D data cubes are formed by 3-manifolds, which can be described and distinguished by using methods of algebraic and geometric topology~\citep{Hatcher2002}.
        The topology of objects in 4D can be much richer than in 2D or 3D and the topological classification of 3-manifolds was only achieved in 2003~\citep{BessieresEtAl2010}. In the present study, we only consider some basic 4D objects as part of our dataset generation, namely balls, 
        that is, $B^4 = \{ x \in \mathbb{R}^4; ||x|| \le~1 \}$, and various tori that exist in 4D, including $S^1 \times B^3$, $S^2 \times B^2$, and $S^1 \times S^1 \times B^2$.
        However, these manifolds are already topologically more complex than what would usually be considered in machine learning, for example, as the outcome of manifold learning~\citep{LeeVerleysenBook2007}.
 
        The Betti numbers are a concept in algebraic topology that captures the essential structure of a manifold or topological space given by the holes of the manifold or topological space~\citep{Ede10}. The $k$\textsuperscript{th} Betti number is often denoted by $\beta_k$, where $\beta_0$ is the number of path-connected components that comprise a topological space, and $\beta_{k \ge 1}$ are the number of $k$-dimensional holes in the space. 
        Holes are formalised in algebraic topology, where roughly speaking, a $k$-dimensional \emph{cycle} is a closed submanifold, a $k$-dimensional \emph{boundary} is a cycle that is also the boundary of a submanifold, and a $k$-dimensional \emph{homology class} is an equivalence class of the group of cycles modulo the group of boundaries $Z_k/B_k$, otherwise known as the $k$\textsuperscript{th} homology group $H_k$. Any non-trivial homology class represents a cycle that is not a boundary, or equivalently, a $k$-dimensional hole. $\beta_k$ can be defined as the rank of the group $H_k$~\citep{Ede10}.
        In this work, the term \textit{hole} will be used in its homological sense, and the term \textit{cavity} will refer to the result of `cutting-out' of the interior of a sample. For example, the introduction of a donut-shaped cavity into a 3D sample, that is $I^3-S^1 \times B^2$, will result in the introduction of a 1D hole and a 2D hole.
        
        In $\mathbb{R}^4$, only the first four Betti numbers are relevant, where $\beta_0$ corresponds to the number of connected components, $\beta_1$ corresponds to the number of circular holes, $\beta_2$ counts the number of 3D voids or tunnels, that is 2D holes, and $\beta_3$ indicates how often the encapsulation of a 4D space occurs. 
        The Betti numbers are a more fine-grained signature than the more broadly known Euler characteristic $\chi$. Their relationship  is given by 
        \begin{equation}
            \chi = \sum_{k=0}^\infty(-1)^k\beta_k.
        \end{equation}
        $\chi$ can be defined in several ways, where the above equation is one option ~\citep{Adhikari2022,tomDieck2008}.
        
        Persistent homology is a computational approach with which one can derive topological indices, such as the Betti numbers, of the underlying manifold of data. 
        The theoretical complexity of applying persistent homology using a sparse implementation is cubic in the number of simplices that describe a sample, however, in practice, this can be as low as linear~\citep{Zom05}.
        A general introduction that can serve as background to computational topology and algebraic topology can be found in the books of~\citet{Ede10} and~\citet{Hatcher2002}, respectively.   
        
        The use of CNNs to predict the Betti numbers of data was first proposed by~\citet{Pau19}, who conducted supervised training of 2D and 3D CNNs using simulated data cubes into which cavities were introduced and labelled with Betti numbers. This approach was extended in a previous pilot study~\citep{Han23},
        which used simulated 4D data cubes with Betti number labels to train a custom 4D CNN that was implemented using the Pytorch library~\citep{Pas19}. 
        Both studies used persistent homology software (JavaPlex~\citep{AdamsEtAl2014} and GUDHI~\citep{Mar14}, respectively) as a comparison partner, and discussed some of the headwinds that were faced when using both persistent homology software and CNNs. 
        When analysing image-type data with persistent homology, it was possible to gain some memory and speed advantage by using a single (unfiltered) cubical complex~
        \citep{Ott16, Rob11, Del15}.
        Notwithstanding, these headwinds appear to magnify in the 4D setting, where the computational and memory demands of analysing samples larger than $64^4$ became prohibitively large, even when aided by basic supercomputers available at that time.

        Persistent homology algorithms are often used to summarise some of the topological and geometrical attributes of a dataset by distilling them into a visual output, and there is evidence that subsampling methods can be effective when used to compute averaged persistence images~\citep{Cha15b}, diagrams~\citep{Cao22} and landscapes~\citep{Sol22} of point-cloud data. \citet{Moi18} proposed a clustering approach to facilitate the persistent homology algorithm, and~\citet{Nan22} explored sampling techniques in the multi-parameter context.
        In the present study, we consider standard downsampling and average-pooling techniques to downscale large 4D image-type manifold data, and demonstrate that while persistent homology algorithms may begin to break down when analysing the global topology of these downscaled data (that is, they may compute results that are vastly different from those attributed to the original data), CNNs appear to better tolerate the use of these techniques as a means to mitigate the limitations that are faced when estimating the Betti numbers of these data.
        While more sophisticated downscaling approaches do exist in the 2D image setting, such as content-adaptive~\citep{Kop13}, perceptually-based~\citep{Ozt15}, and detail-preserving~\citep{Web16} algorithms,
        the techniques considered in this work offer an early look into how pre-processing 4D image-type data may afford the analysis of larger samples, with potentially higher resolutions, or a greater number~of~cavities.
        As we will discuss later in Section~\ref{sec:discussion}, our results also serve to motivate an investigation into the use of these other algorithms, along with other machine learning approaches, as they may complement the results that are presented here.

        \begin{table*}[ht!]
            \caption{Describing 4D cavities in an $(x,y,z,w)$-system}
            \label{tab:implicit_formulas}
            \centering
                \begin{tabular}{cccc}
                    \toprule
                    Manifold & Formula & Volume & Simplification \\[1mm]
        
                    \midrule
                    $B^4$ & $x^2 + y^2 + z^2 + w^2 \le a^2$ & $\frac{\pi^2}{2}a^4$ & \\[1mm]
                    
                    \midrule
                    $S^1 \times B^3$ & $( \sqrt{x^2 + y^2} -R )^2 + z^2 + w^2 \le a^2$ & $\frac{8}{3} \pi^2 R a^3$ & $\frac{16}{3} \pi^2 a^4$ \\[1mm]
                    
                    \midrule
                    $S^2 \times B^2$ & $(\sqrt{x^2 + y^2 + z^2} - R)^2 + w^2 \le a^2$ & $4 \pi^2 R^2 a^2$ & $16 \pi^2 a^4$ \\[1mm]
        
                    \midrule
                    \multirow{3}{*}{$S^1 \times S^1 \times B^2$} & \multicolumn{1}{l}{$( \sqrt{(B(\sqrt{x^2 + y^2} - R) - Aw)^2 + z^2} - r )^2 +$ } & \multirow{3}{*}{$4 \pi^3 R r a^2$} & \multirow{3}{*}{$16 \pi^3 a^4$ to $32 \pi^3 a^4$} \\
                    & \multicolumn{1}{c}{$( A(\sqrt{x^2 + y^2} - R) + Bw )^2 \le a^2,$} &&\\
                    & \multicolumn{1}{r}{where $A=\cos{\alpha}$ and $B=\sin{\alpha}$} &&\\[1mm]
                    
                    \bottomrule
                \end{tabular}
        \end{table*}	

    \section{4D Dataset generation}\label{sec:dataset_generation}
        
        The supervised training approach of our study uses synthesised 4D data cubes with topological labels. Data generation software was implemented in Python using data structures from the NumPy library~\citep{Har20} to represent 4D data cubes as images, and apply vector and matrix operations. Beginning with a `solid' 4D cube (represented by a 4D $128^4$ tensor with every entry set to 1), a random number of cavities were introduced into the cube by setting the entries that represented the cavities to~0. 
        Each cavity was homeomorphic to one of the objects in Table~\ref{tab:implicit_formulas}, and was randomly scaled and rotated before being positioned.
        The Betti numbers (chosen according to match a possible combination of cavities) and degree of cavity scaling were uniformly distributed.
        The resulting cube was a 4D generalisation of a single-channelled, black-and-white image.
        
        \subsection{Design}\label{subsec:design}
            Data design experiments were focused on choosing radius parameters (namely, $a$, $r$, and $R$) that would generate samples with non-trivial topologies in a resolution that would ensure that holes were represented clearly (in a homological sense). 
            Since the toxels of an image were attributed integral coordinates, persistent homology software would, theoretically, be capable of correctly detecting holes of any dimension, provided that the diameter of a hole was greater than the distance between diagonal points of a 4D unit-cube ($\sqrt{4}$ units). Otherwise, calculations would suffer as a result of there being insufficient resolution to describe a hole with such a small radius. Conversely, if parameters were too large, then the cavities would be so big that we would be limited to samples with fewer holes and less interesting topologies.
            
            Table~\ref{tab:implicit_formulas} provides several formulas that may be used to depict a variety of 4D objects in an $(x,y,z,w)$-system and were used to design the cavities for the dataset.
            The formulas were found by firstly compressing trigonometrically-derived parametric equations into implicit formulas, and then replacing the equality in each formula with an inequality in order to describe a `solid' object that could be removed from the interior of a 4D cube; further detail and an example can be found in the appendices of the workshop paper by~\citet{Han23}.
            
            These objects vary in their geometry (relative to each other), and this can be demonstrated by experimenting with the parameters in each formula. For example, $B^4$ does not have a tunnel, whereas $S^1 \times S^1 \times B^2$ does have a tunnel and can vary from being quite `flat and expansive' to being more `round', depending on the choice of the parameter $\alpha$, which sets the orientation of the $S^1 \times B^2$ factor and ranges from $0$ to $\pi/2$. 
            Figure~\ref{fig:visualising_S1S1S1} shows several 2D visualisations that offer some intuition of how varying $\alpha$ can impact the resulting embedding of $S^1 \times S^1 \times S^1$ (and, equivalently, $S^1 \times S^1 \times B^2$) in $\mathbb{R}^4$. The idea is to begin with a (dark-grey) torus $S^1 \times S^1$ that is oriented according to $\alpha$, and positioned $R$ units from the origin along the $x$-axis in the 3D $xzw$-hyperplane. The torus is then rotated around the origin, through the $xy$-plane, in order to introduce the third $S^1$ factor. Although the figures may suggest otherwise, the implicit formula for $S^1 \times S^1 \times B^2$ that is used to generate our data guarantees that overlaps or self-intersections do not occur. This is because we are working in $\mathbb{R}^4$, rather than $\mathbb{R}^3$, as the figures may also suggest; the extra dimension cannot be shown easily in 2D.
            Figure~\ref{fig:embedding_S1S1B2} demonstrates a construction in which $\alpha = 0$, and Figures~\ref{fig:embedding_alt_S1S1B2} and \ref{fig:embedding_rot_alt_S1S1B2} demonstrate constructions in which $\alpha = \pi/2$; notice that Figure~\ref{fig:embedding_rot_alt_S1S1B2} is, in fact, a $\pi/2$ radians $zw$-rotation of Figure~\ref{fig:embedding_alt_S1S1B2}. Because of the symmetry of the torus that we begin with, any $zw$-rotation of the construction in Figure~\ref{fig:embedding_S1S1B2} is inconsequential, as is a $\pi$ radians $zw$-rotation of the remaining examples. In practice, we only need to consider when $\alpha$ ranges from $0$ to $\pi/2$ because the remaining angles arise freely from the random rotations that are applied during data generation.
            
            \tikzset{
              pics/torus/.style n args={3}{
                code = {
                  \providecolor{pgffillcolor}{rgb}{1,1,1}
                  \begin{scope}[
                      yscale=cos(#3),
                      outer torus/.style = {draw=black,line width/.expanded={\the\dimexpr2\pgflinewidth+#2*2},line join=round},
                      inner torus/.style = {draw=pgffillcolor,line width={#2*2}}
                    ]
                    \draw[outer torus] circle(#1);
                    \draw[inner torus] circle(#1);
                    
                    \ifnum #3>180
                    {
                        \draw[outer torus] (180:#1) arc (180:360:#1);
                        \draw[inner torus,line cap=round] (180:#1) arc (180:360:#1);
                    }
                    \else
                    {
                        \ifnum #3=180
                        {
                            \draw[outer torus] (0:#1) arc (0:180:#1); 
                            \draw[inner torus,line cap=round] (0:#1) arc (0:360:#1);
                        }
                        \else
                        {   
                            \ifnum #3=0
                            {
                                \draw[outer torus] (0:#1) arc (0:180:#1); 
                                \draw[inner torus,line cap=round] (0:#1) arc (0:360:#1);
                            }
                            \else
                            {
                                \draw[outer torus] (0:#1) arc (0:180:#1);
                                \draw[inner torus,line cap=round] (0:#1) arc (0:180:#1);
                            }
                            \fi
                        }
                        \fi
                    }
                    \fi
            
                  \end{scope}
                }
              }
            }
            
            \tikzset{
              pics/torus_flat/.style n args={3}{
                code = {
                  \providecolor{pgffillcolor}{rgb}{1,1,1}
                  \begin{scope}[
                      yscale=cos(60),
                      outer torus/.style = {draw=black,line width/.expanded={\the\dimexpr2\pgflinewidth+#2*2},line join=round},
                      inner torus/.style = {draw=pgffillcolor,line width={#2*2}}
                    ]
                    \draw[outer torus] circle(#1);
                    \draw[inner torus] circle(#1);
                    
                    \draw[outer torus] (0:#1) arc (0:180:#1);
                    \draw[inner torus,line cap=round] (0:#1) arc (0:180:#1);

                  \end{scope}
                }
              }
            }

            \begin{figure}[ht!]
                \centering
                \begin{subfloat}[$\alpha = 0$] 
                {
                    \centering
                    \begin{tikzpicture}
                        \draw [-to,line width=0.5pt] (0,0)-- (0,-1.5);
                        
                        \foreach \i in{180,148,125,111,100}
                            \pic[fill=lightgray!\i!lightgray, rotate=90] at ({\i-90}:{0.9/(1-(0.9*cos(\i-90))^2 )^0.5} ) {torus={2mm}{1mm}{\i} };
    
                        \foreach \i in{90}
                            \pic[fill=gray!\i!gray, rotate=90] at ({\i-90}:{0.9/(1-(0.9*cos(\i-90))^2 )^0.5} ) {torus={2mm}{1mm}{\i} };
    
                        \foreach \i in{80,69,55,32}
                            \pic[fill=lightgray!\i!lightgray, rotate=90] at ({\i-90}:{0.9/(1-(0.9*cos(\i-90))^2 )^0.5} ) {torus={2mm}{1mm}{\i} };
                        
                        \foreach \i in{212,235,249,260,270,280,291,305,328,0}
                            \pic[fill=lightgray!\i!lightgray, rotate=90] at ({\i-90}:{0.9/(1-(0.9*cos(\i-90))^2 )^0.5} ) {torus={2mm}{1mm}{\i} };
                        
                        \draw [-to,line width=0.5pt] (0,0)-- (0,1.5);
                     \end{tikzpicture}
                     \label{fig:embedding_S1S1B2}
                }
                \end{subfloat}
                \hfill
                \begin{subfloat}[$\alpha = \pi/2$]
                {
                    \centering
                    \begin{tikzpicture}
                        \draw [-to,line width=0.5pt] (0,0)-- (0,-1.5);
                        
                        \foreach \i in{90,70,50,30,15}
                            \pic[fill=lightgray!\i!lightgray, rotate=90] at (\i:{0.9/(1-(0.9*cos(\i))^2 )^0.5} ) {torus={2mm}{1mm}{\i} };
    
                        \foreach \i in{0}
                            \pic[fill=gray!\i!gray, rotate=90] at (\i:{0.9/(1-(0.9*cos(\i))^2 )^0.5} ) {torus={2mm}{1mm}{\i} };
    
                        \foreach \i in{345,330,310,290,270}
                            \pic[fill=lightgray!\i!lightgray, rotate=90] at (\i:{0.9/(1-(0.9*cos(\i))^2 )^0.5} ) {torus={2mm}{1mm}{\i} };
                        
                        \foreach \i in{110,130,150,165,180,195,210,230,250}
                            \pic[fill=lightgray!\i!lightgray, rotate=90] at (\i:{0.9/(1-(0.9*cos(\i))^2 )^0.5} ) {torus={2mm}{1mm}{\i} };
                        
                        \draw [-to,line width=0.5pt] (0,0)-- (0,1.5);
                     \end{tikzpicture}
                     \label{fig:embedding_alt_S1S1B2}
                }
                \end{subfloat}
                \hfill
                \begin{subfloat}[$\alpha = \pi/2$ with $zw$-rotation]
            {
                \centering
                \begin{tikzpicture}
                    \draw [-to,line width=0.5pt] (0,0)-- (0,-1.5);
                    
                    \foreach \i in{90,60,35,20,10,0}
                        \pic[fill=lightgray!\i!lightgray, rotate=180] at (\i:{0.9/(1-(0.9*cos(\i))^2 )^0.5} ) {torus_flat={2mm}{1mm}{\i} };

                    \foreach \i in{0}
                        \pic[fill=gray!\i!gray, rotate=180] at (\i:{0.9/(1-(0.9*cos(\i))^2 )^0.5} ) {torus_flat={2mm}{1mm}{\i} };

                    \foreach \i in{350,340,325,300}
                        \pic[fill=lightgray!\i!lightgray, rotate=180] at (\i:{0.9/(1-(0.9*cos(\i))^2 )^0.5} ) {torus_flat={2mm}{1mm}{\i} };
                    
                    \foreach \i in{120,145,160,170,180,190,200,215,240,270}
                        \pic[fill=lightgray!\i!lightgray, rotate=180] at (\i:{0.9/(1-(0.9*cos(\i))^2 )^0.5} ) {torus_flat={2mm}{1mm}{\i} };
                    
                    \draw [-to,line width=0.5pt] (0,0)-- (0,1.5);
                 \end{tikzpicture}
                 \label{fig:embedding_rot_alt_S1S1B2}
            }
            \end{subfloat}

                \caption{Graphical visualisations demonstrating the impact of varying $\alpha$ on $S^1 \times S^1 \times S^1$.
                The vertical double arrows represent the $z$-axis.
                The dark-grey tori are oriented according to $\alpha$, and positioned $R$ units from the origin along the $x$-axis in the 3D $xzw$-hyperplane. We can then imagine rotating these tori through the $xy$-plane in order to produce each figure. The vertical proportions of each torus sits along the $z$-axis, the horizontal proportions sit along the $w$-axis, and the radial proportions sit along the respective (radial) vector in the $xy$-plane that is directed from the origin towards each torus.
                In (a), $\alpha$ is set to 0, and in (b) and (c), $\alpha$ is set to $\pi/2$. }
                \Description{Three examples of how S1 times S1 times S1 can be visualised.}
        		\label{fig:visualising_S1S1S1}
            \end{figure}

        \subsection{Hypervolumes}\label{subsec:hypervolumes}
            In order to maintain some homogeneity in the range of sizes of the objects that we considered, we selected parameters for each object that would produce cavities with a similar range of hypervolume (4D-volume). Formulas for the hypervolumes of these objects, along with some simplifications that arise by setting $R=2a$ (for $S^1 \times B^3$ and $S^2 \times B^2$) are also provided in Table~\ref{tab:implicit_formulas}. For $S^1 \times S^1 \times B^2$, we assume that $r=2a$, and that the value of $R$ depends on $\alpha$ and ranges from $2a$ to $4a$. Therefore, for $a \in [a_\mathrm{min}, a_\mathrm{max} ]$, the hypervolume of $S^1 \times S^1 \times B^2$ ranged from $16 \pi^3 a_\mathrm{min}^4$ to $32 \pi^3 a_\mathrm{max}^4$.
            
            The hypervolumes of the remaining objects were scaled into the same range by finding suitable values for $a_{object}$. For example, if the hypervolume of $B^4$ were to also fall within this range, it was necessary to choose $a_{B^4}$ such that $\frac{\pi^2}{2}a_{B^4}^4 \in [16 \pi^3 a_\mathrm{min}^4, 32 \pi^3 a_\mathrm{max}^4]$. Rearranging this expression leads to Equation~\ref{eqn:ball_range}. The remaining ranges given in Equations~\ref{eqn:S1xB3_range} and~\ref{eqn:S2xB2_range} were deduced in the same way.
        	\begin{equation}\label{eqn:ball_range}
                a_{B^4} \in [2\sqrt[4]{2 \pi} a_\mathrm{min}, 2\sqrt[4]{4 \pi} a_\mathrm{max}]
        	\end{equation}
        	\begin{equation}\label{eqn:S1xB3_range}
                a_{S^1 \times B^3} \in [\sqrt[4]{3 \pi} a_\mathrm{min}, \sqrt[4]{6 \pi} a_\mathrm{max}]
        	\end{equation}
        	\begin{equation}\label{eqn:S2xB2_range}
                a_{S^2 \times B^2} \in [\sqrt[4]{\pi} a_\mathrm{min}, \sqrt[4]{2 \pi} a_\mathrm{max}]
        	\end{equation}
        
        \subsection{Dataset parameters}\label{subsec:dataset_parameters}
            The parameters that we selected in order to generate a 4D dataset with which to investigate our approach are summarised in Table \ref{tab:dataset_parameters}. 
            Our study focused on recognising the global topology of compact manifolds. Hence, we restricted our experiments to single-component samples ($\beta_0 = 1$), and ensured that both a cube's boundaries were not disturbed and that cavities did not intersect each other. These rules were enforced by implementing a 1-toxel boundary and minimum 6.5-unit spacing between cavities. The self-intersection of a cavity was prevented via the simplifications that were explained in Section~\ref{subsec:hypervolumes}.
            The choices for $a_\mathrm{min}$ and $a_\mathrm{max}$, as listed in the $B^2$ row of the $S^1 \times S^1 \times B^2$ column of Table~\ref{tab:dataset_parameters}, were sufficient to produce cavities in an appropriate resolution for the chosen dimensions of our samples.
            
            \begin{table*}[ht!]
                \caption{The dataset parameters and the unit radius ranges for each factor of the manifolds that were used to produce 4D cavities; $\alpha$ is expressed in radians. }
                \label{tab:dataset_parameters}
                \centering
                    \begin{tabular}{cccccc}
                    \toprule
                        Cube & Parameters & $B^4$ & $S^1 \times B^3$ & $S^2 \times B^2$ & $S^1 \times S^1 \times B^2$ \\
                        \midrule
                        \multirow{4}{*}{ $128^4$ }& 32000 samples & $B^4$: 7.6 to 24.1 & $S^1$: 8.4 to 26.7 & $S^2$: 6.4 to 20.3 & $S^1$: 4.8 to 25.6 \\
                        & 1 toxel boundary && $B^3$: 4.2 to 13.3 & $B^2$: 3.2 to 10.1 & $S^1$: 4.8 to 12.8 \\
                        & 1 to 48 holes &&&& $B^2$: 2.4 to 6.4  \\
                        & 6.5 unit spacing &&&& $\alpha$: 0 to $\pi/2$ \\
                        
                        \bottomrule
                    \end{tabular}
            \end{table*}

            Each sample comprised of a 4D $128^4$ cube, with the combined number of 1D, 2D, and 3D holes ranging from 1 to 48.
            A $128^4$ cube was used because it was large enough to contain a non-trivial range of cavities, which afforded the analysis of samples with interesting topologies. This data was also slightly bigger than what would have been feasible to be directly analysed using persistent homology methods or CNNs with our hardware (NVIDIA DGX Station, with an Intel Xeon E5-2698 v4 CPU, 256GB RAM, and four V100-32GB GPUs). Hence, downscaling became a requirement in order to process this data.     
            Note also that the cavity dimensions were small enough that the application of downscaling could potentially disturb the homology of a sample by closing up holes (see Section~\ref{subsec:downsampling_approach}).
            

            A dataset of 32000 samples was acquired.
            The data was generated in parallel on a High Performance Computing (HPC) Grid in 100-sample batches, over 320 nodes. An average of 8.60 hours was required to complete each batch, and the entire process utilised approximately 2753 HPC hours.
            
\begin{figure*}[ht!]
                \centering
                \includegraphics[trim=0 3cm 0 4cm,clip,width=0.28\textwidth,height=0.18\textwidth]{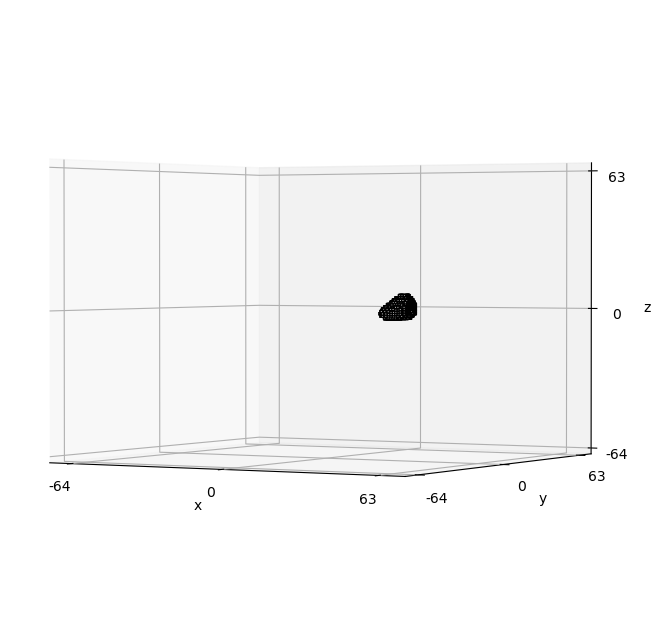}
                \includegraphics[trim=0 3cm 0 4cm,clip,width=0.28\textwidth,height=0.18\textwidth]{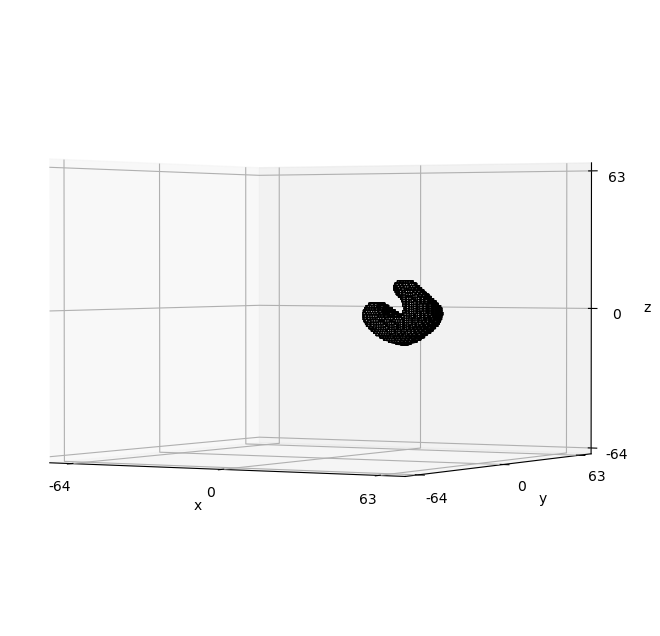}
                \includegraphics[trim=0 3cm 0 4cm,clip,width=0.28\textwidth,height=0.18\textwidth]{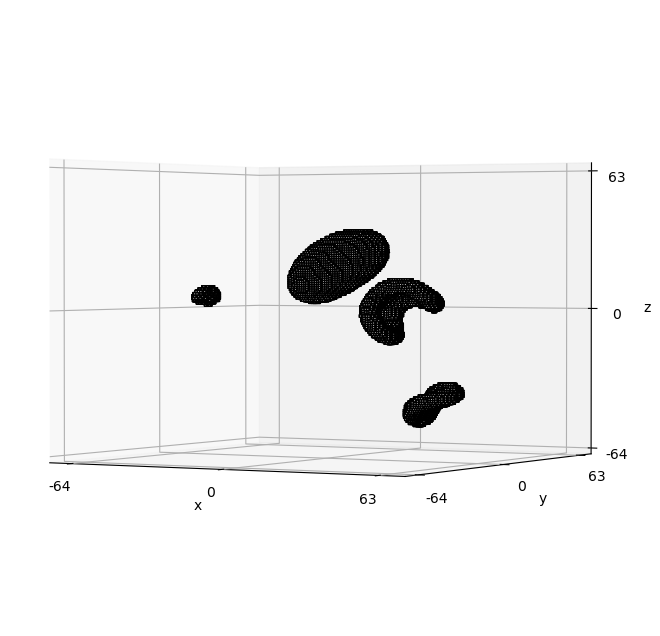}
                \includegraphics[trim=0 3cm 0 4cm,clip,width=0.28\textwidth,height=0.18\textwidth]{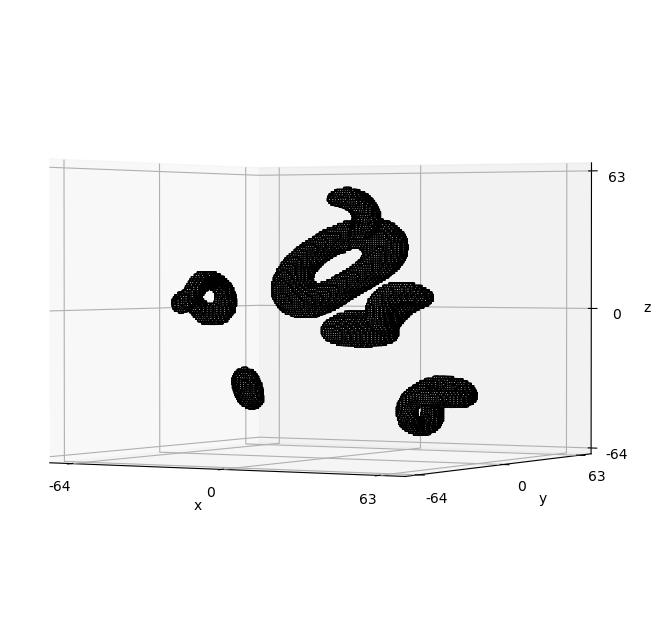}
                \includegraphics[trim=0 3cm 0 4cm,clip,width=0.28\textwidth,height=0.18\textwidth]{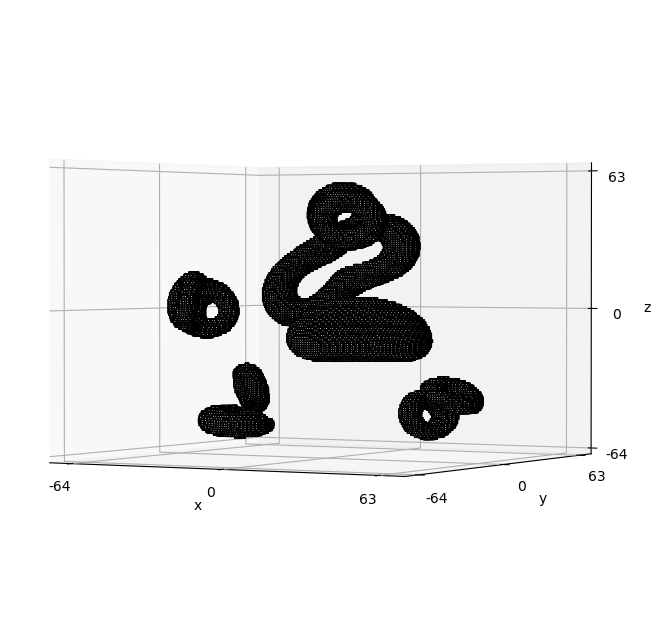}
                \includegraphics[trim=0 3cm 0 4cm,clip,width=0.28\textwidth,height=0.18\textwidth]{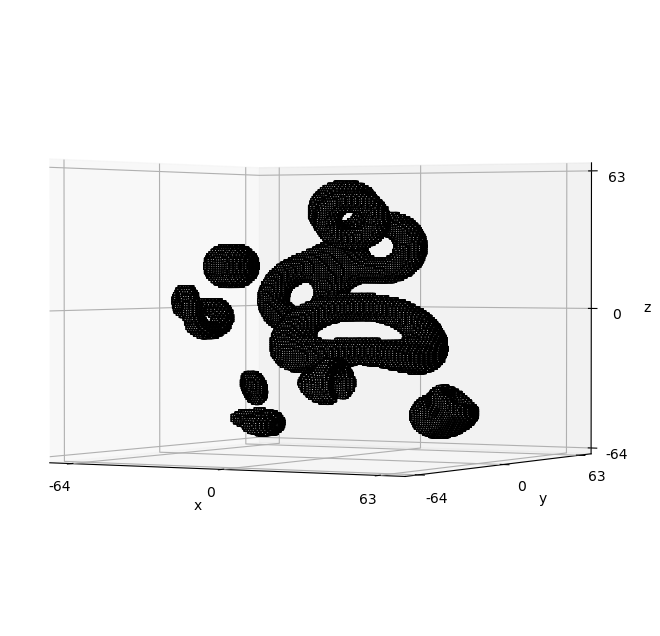}
                \includegraphics[trim=0 3cm 0 4cm,clip,width=0.28\textwidth,height=0.18\textwidth]{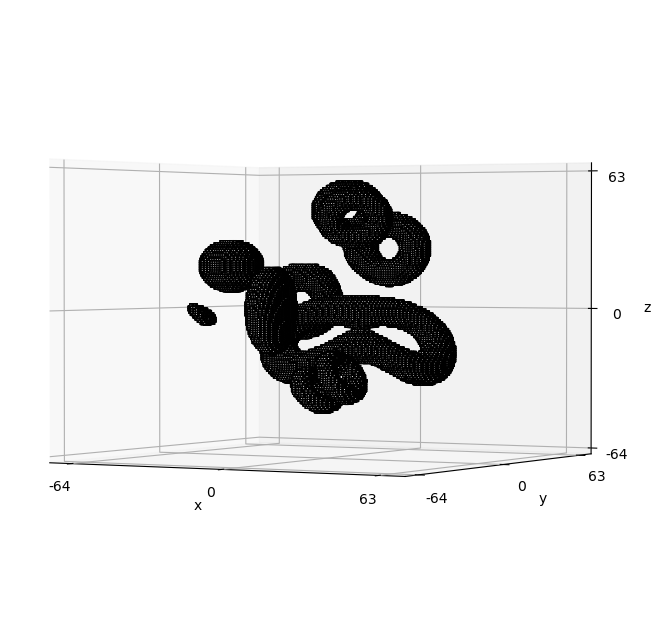}
                \includegraphics[trim=0 3cm 0 4cm,clip,width=0.28\textwidth,height=0.18\textwidth]{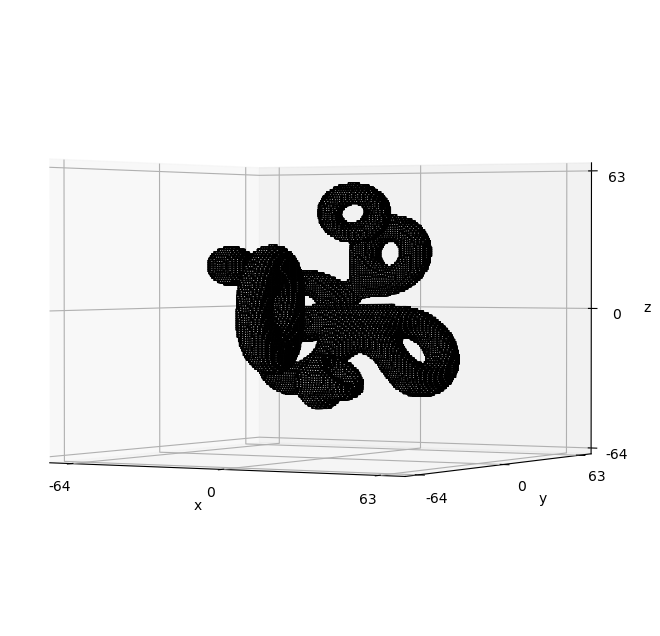}
                \includegraphics[trim=0 3cm 0 4cm,clip,width=0.28\textwidth,height=0.18\textwidth]{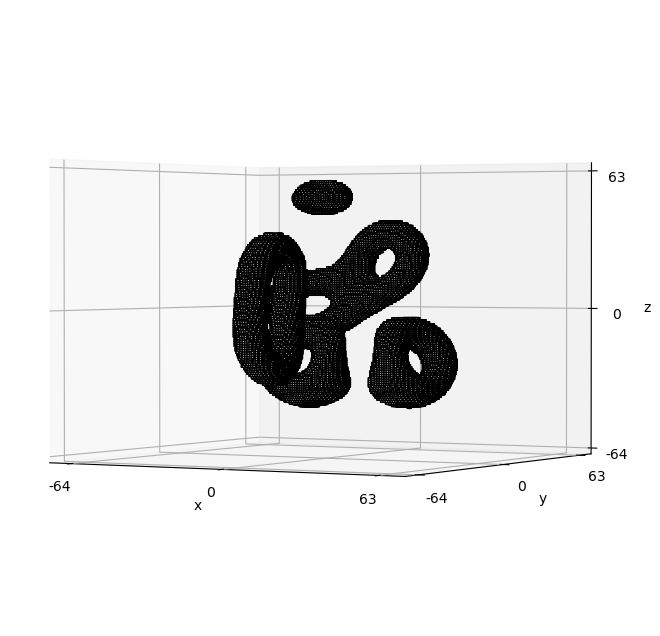}
                \includegraphics[trim=0 3cm 0 4cm,clip,width=0.28\textwidth,height=0.18\textwidth]{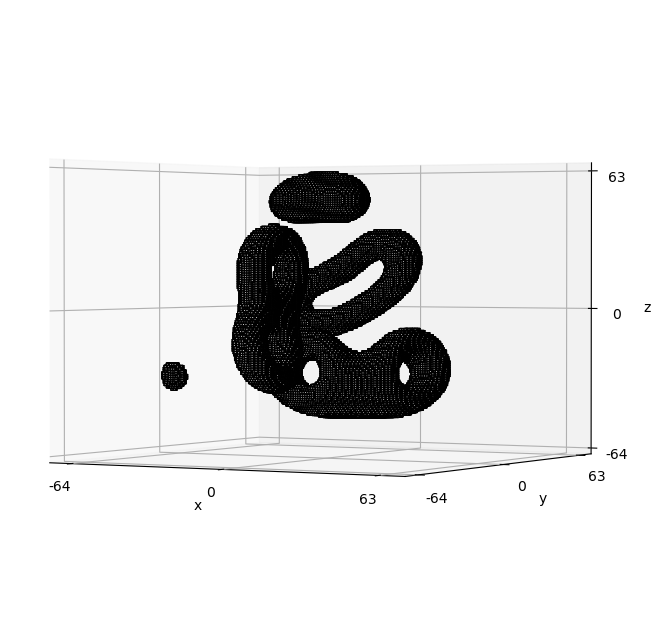}
                \includegraphics[trim=0 3cm 0 4cm,clip,width=0.28\textwidth,height=0.18\textwidth]{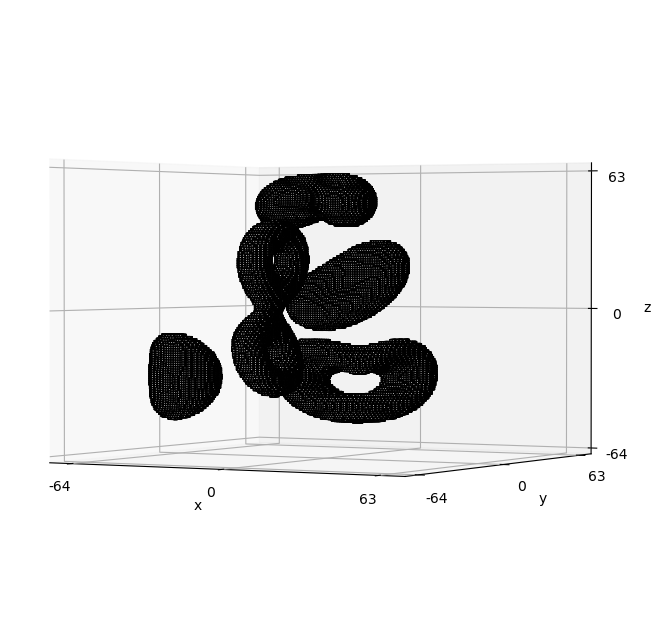}
                \includegraphics[trim=0 3cm 0 4cm,clip,width=0.28\textwidth,height=0.18\textwidth]{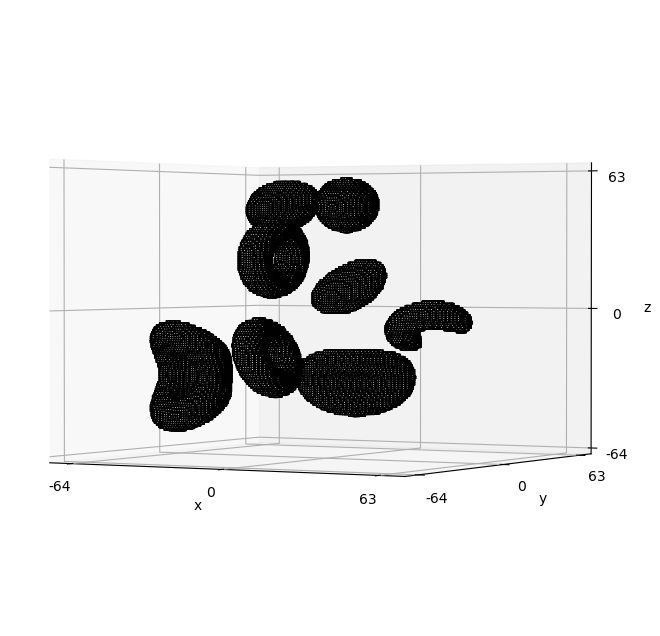}
                \includegraphics[trim=0 3cm 0 4cm,clip,width=0.28\textwidth,height=0.18\textwidth]{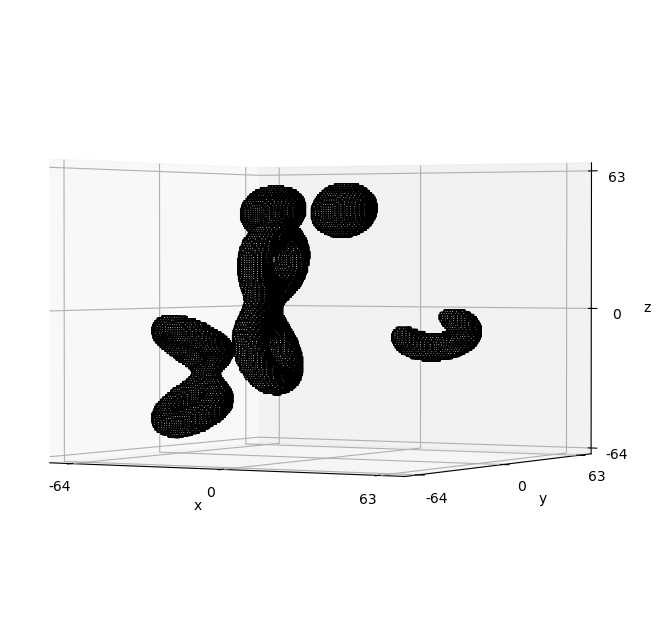}
                \includegraphics[trim=0 3cm 0 4cm,clip,width=0.28\textwidth,height=0.18\textwidth]{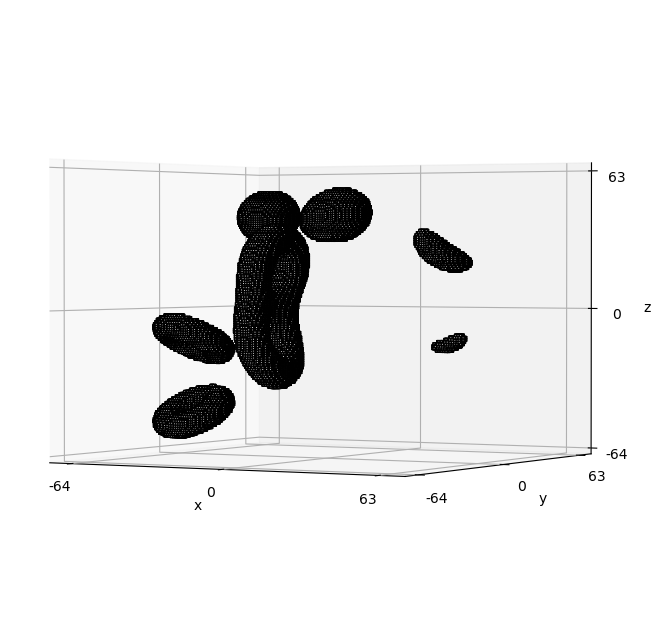}
                \includegraphics[trim=0 3cm 0 4cm,clip,width=0.28\textwidth,height=0.18\textwidth]{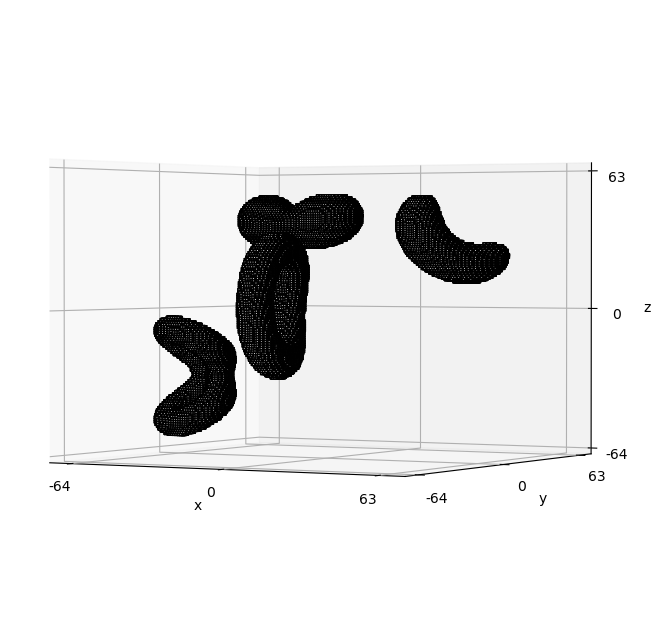}
                \includegraphics[trim=0 3cm 0 4cm,clip,width=0.28\textwidth,height=0.18\textwidth]{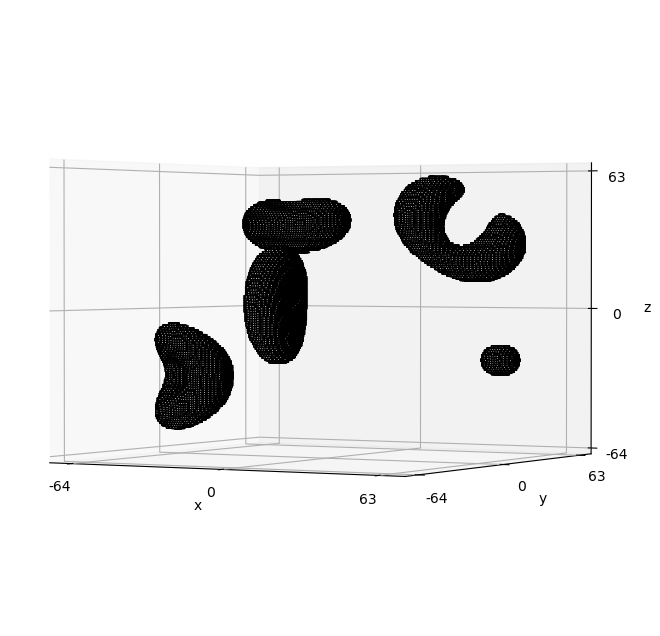}
                \includegraphics[trim=0 3cm 0 4cm,clip,width=0.28\textwidth,height=0.18\textwidth]{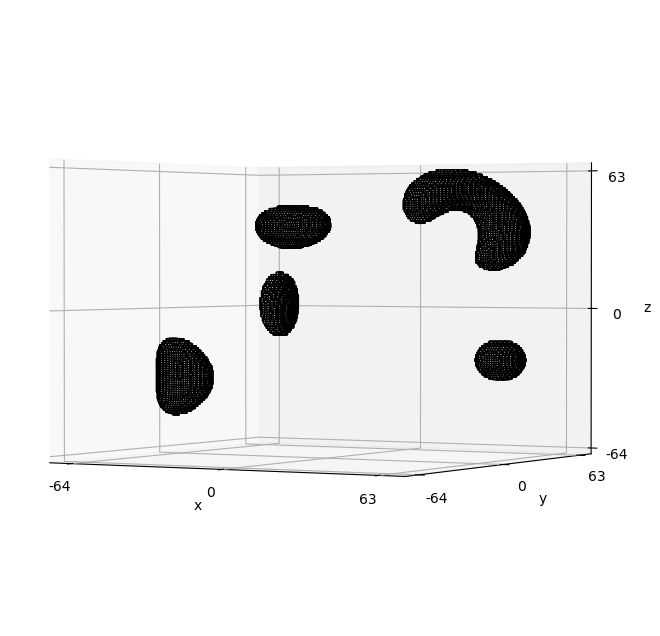}
                \includegraphics[trim=0 3cm 0 4cm,clip,width=0.28\textwidth,height=0.18\textwidth]{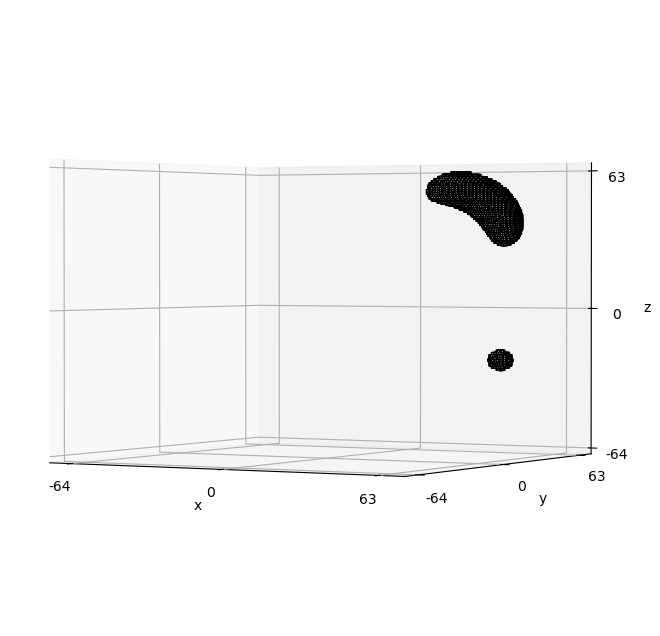}
                    
                \caption{A $128^4$ sample is inspected by taking 18 equally-spaced 3D slices along the $w$-axis. The slices are ordered from left to right and top to bottom. The toxel values are inverted (setting 0 to 1, and vice versa) in order to reveal the cavities within a sample. This sample was created by introducing two $B^4$ cavities, four $S^1 \times B^3$ cavities, one $S^2 \times B^2$ cavity, and nine $S^1 \times S^1 \times B^2$ cavities into a 4D cube. For example, a 4-ball is seen in the bottom right corner of the last three slices and an example of $S^1 \times B^3$ is seen at the top centre of slices nine to seventeen as an object that splits into two and then merges back into one.}
                \Description{Eighteen equally-spaced 3D slices from a 4D sample that contains a variety of cavities.}
        		\label{fig:visualising_a_4d_sample}
            \end{figure*}

            A visualisation of the cavities within a $128^4$ sample is shown in Figure~\ref{fig:visualising_a_4d_sample}. This is achieved by inverting the toxel values (setting 0 to 1, and vice versa) so that the cube itself is stripped away in order to reveal the objects that were used to produce its cavities, and then taking 3D slices along some axis~\citep{Pre84}. In this case, 18 equally-spaced slices have been taken along the $w$-axis. Taking finer slices allows one to see a more continuous-looking evolution of the cavities (see Appendix~\ref{app_subsec:visualising_4d_samples}). 

        \subsection{Data labelling}\label{subsec:data_labelling}
            Each label was produced on-the-fly during the generation of a sample. 
            The homology of each cavity that was introduced into a sample was algebraically derived by, firstly, observing that the 4D cube $I^4$ is homeomorphic to the 4D ball and therefore shares the same homology. Secondly, the homology of both the object being considered for removal $M$ and its boundary $\partial M$ were computed by using the K\"unneth theorem. Thirdly, the Mayer-Vietoris Sequence was applied to find the homology of the cube with the cavity $I^4 - M$; an introduction to the theorems that were used in this derivation can be found in~\citep{Hatcher2002}. The results of these calculations are provided in Table~\ref{tab:betti_numbers}; the manifolds involving a subtraction from $I^4$ were the most relevant to this work. A more detailed description of the mathematics involved in producing these results can be obtained from a previous workshop paper~\citep{Han23},
            although, the key ideas are summarised in Appendix~\ref{app_sec:algebraic_topology} and example derivations are provided in Appendix~\ref{app_subsec:label_derivation}.

            The primary label of each sample was a vector of Betti numbers, which took the form $[\beta_0, \beta_1, \beta_2, \beta_3]$. A secondary label was also included, which encoded the number of times that each manifold had been removed from the original 4D cube by using a vector that was ordered $[B^4, S^1 \times B^3, S^2 \times B^2, S^1 \times S^1 \times B^2]$. 
            Since the Betti numbers that each cavity contributed to a sample were known, the Betti numbers could simply be summed over their dimensions in order to produce a label for the sample. Furthermore, the parameters that were outlined in Table~\ref{tab:dataset_parameters} were large enough to always produce a homologically correct representation in the $128^4$ cube setting.
            Consequently, there was no requirement to analyse the samples using persistent homology software in order to produce a label.
            
            For example, the sample that is presented in Figure~\ref{fig:visualising_a_4d_sample} was generated by introducing two $B^4$ cavities, four $S^1 \times B^3$ cavities, one $S^2 \times B^2$ cavity, and nine $S^1 \times S^1 \times B^2$ cavities into a 4D cube. By construction, $\beta_0$ is 1. The remaining Betti numbers are found by arranging the first, second and third Betti numbers of $I^4-B^4$, $I^4-(S^1 \times B^3)$, $I^4-(S^2 \times B^2)$, and $I^4-(S^1 \times S^1 \times B^2)$, as vectors of the form $[\beta_1, \beta_2, \beta_3]$, so that we have
            $[0,0,1]$, $[0,1,1]$, $[1,0,1]$, and $[1,2,1]$, respectively. These vectors are then multiplied by the number of instances of each cavity to find the non-zero Betti numbers, as shown in Equation~\ref{eqn:sample_betti_numbers}.
            \begin{equation}\label{eqn:sample_betti_numbers}
                2[0,0,1] + 4[0,1,1] + [1,0,1] + 9[1,2,1] = [10,22,16]
            \end{equation} 
            That is, $\beta_1 = 10$, $\beta_2 = 22$, and $\beta_3 = 16$. The sample's primary label is then encoded as $[1,10,22,16]$.


            \begin{table}[ht!]
                \caption{Betti numbers $\beta_i$ and the Euler characteristic $\chi$ of selected low-dimensional manifolds. The manifolds involving a subtraction from $I^4$ were the most relevant to this work.}
                \label{tab:betti_numbers}

                \centering
                \begin{tabular}{lccccc}
                    \toprule 
                    \multicolumn{1}{c}{Manifold} &$\beta_0$&$\beta_1$&$\beta_2$&$\beta_3$&$\chi$\\
                    \midrule

                    3-Sphere $S^3$&1&0&0&1&0\\
                    4-Ball $B^4$&1&0&0&0&1\\
                    $I^4-B^4$&1&0&0&1&0\\
                    
                    \midrule
                    $S^1 \times S^2$&1&1&1&1&0\\
                    $S^1 \times B^3$&1&1&0&0&0\\
                    $S^2  \times B^2$&1&0&1&0&2\\
                    $I^4-(S^1 \times B^3)$&1&0&1&1&1\\
                    $I^4-(S^2 \times B^2)$&1&1&0&1&-1\\

                    \midrule
                    $S^1 \times S^1 \times S^1$&1&3&3&1&0\\
                    $S^1 \times S^1 \times B^2$&1&2&1&0&0\\
                    $I^4-(S^1 \times S^1 \times B^2)$&1&1&2&1&1\\
                    
                    \bottomrule
                \end{tabular}
            \end{table}

        \subsection{4D dataset distribution}
            The randomly selected aspects of the data generation process, such as the choice of objects, radii, and Betti number combinations, were uniformly distributed. The explicit distribution of various aspects of the acquired 4D dataset are summarised in Tables~\ref{tab:4d_dataset_dist_betti} and~\ref{tab:4d_dataset_dist_objects}. The statistics demonstrate that all samples comprise of a single homological component ($\beta_0 = 1$). They also show that the instances of cavities appear to be evenly distributed among the samples and not all samples contain an example of each type of cavity.

            \begin{table*}[ht!]
                
                \caption{The percentage of samples with a respective Betti number label. The value of the Betti numbers ranged from 0 to 16.}
                \label{tab:4d_dataset_dist_betti}
                \centering
                \begin{adjustbox}{max width=\textwidth}
                \begin{tabular}{lccccccccccccccccc}
                    \toprule 
                    $\beta_n$ & 0 & 1 & 2 & 3 & 4 & 5 & 6 & 7 & 8 & 9 & 10 & 11 & 12 & 13 & 14 & 15 & 16  \\
                    \midrule
                    
                    0 & 0 & 100 & 0 & 0 & 0 & 0 & 0 & 0 & 0 & 0 & 0 & 0 & 0 & 0 & 0 & 0 & 0 \\

                    1 & 3.53 & 6.55 & 8.82 & 9.84 & 10.49 & 10.45 & 10.16 & 9.18 &  8.17 & 6.46 & 5.4 & 4.14 & 2.69 & 2.19 & 1.12 & 0.58 & 0.22  \\

                    2 & 3.73 & 3.12 & 6.15 & 5.42 & 7.44 & 6.83 & 8.19 & 6.76 & 7.8  & 6.78 & 7.59 & 6.07 & 6.32 & 5.11 & 4.91 & 3.78 & 3.99 \\

                    3 & 0 & 0.12 & 0.24 & 0.5 & 0.81 & 1.31 & 1.94 & 2.9 & 4.09 & 5.15 & 6.33 &  8.37 & 10.02 & 11.52 & 13.56 & 15.45 & 17.67 \\
                    
                    \bottomrule
                \end{tabular}
                \end{adjustbox}
                
            \end{table*}

            \begin{table*}[ht!]
                
                \caption{The number of samples that contain a given cavity and the number of instances that each cavity appears in the dataset.}
                \label{tab:4d_dataset_dist_objects}

                \centering
                \begin{tabular}{lcccc}
                    \toprule 
                    Manifold & $B^4$ & $S^1 \times B^3$ & $S^2 \times B^2$ & $S^1 \times S^1 \times B^2$ \\
                    \midrule
                    
                    Number of samples with at least one instance & 26483 & 25469 & 26521 & 24631 \\
                    Number of instances that appear in dataset & 112720 & 101001 & 111302 & 75864 \\
                    
                    \bottomrule
                \end{tabular}
                
            \end{table*}

    \section{Experiments and results}\label{sec:experiments_and_results_4d}
        The goal of this work was to explore whether downscaling techniques could be useful in circumventing the complexity issues that make the application of persistent homology and machine learning software difficult when analysing the topological characteristics of large 4D image-type manifold data.
        Two downscaling approaches were considered: downsampling, and average-pooling. The GUDHI persistent homology Python library was used to determine how consistent the sample labels were with their downsampled image; a cubical complex was used because of its suitability to image-type data~\citep{Ott16, Rob11, Del15}. The persistent homology algorithm operated over a single cubical complex, which comprised only of cubical simplexes between contiguous voxels with a value equal to 1, that is, the cubical complex was completely determined by the voxels themselves. Therefore, in theory, any persistent homology software would have applied the algorithm to the same complex and yielded the same result.
        
        \subsection{Downsampling approach}\label{subsec:downsampling_approach}
\begin{figure}[ht!]
                \centering
                \begin{minipage}[b]{0.45\textwidth}
                    \centering
                    \includegraphics[width=0.95\textwidth,height=0.7\textwidth]{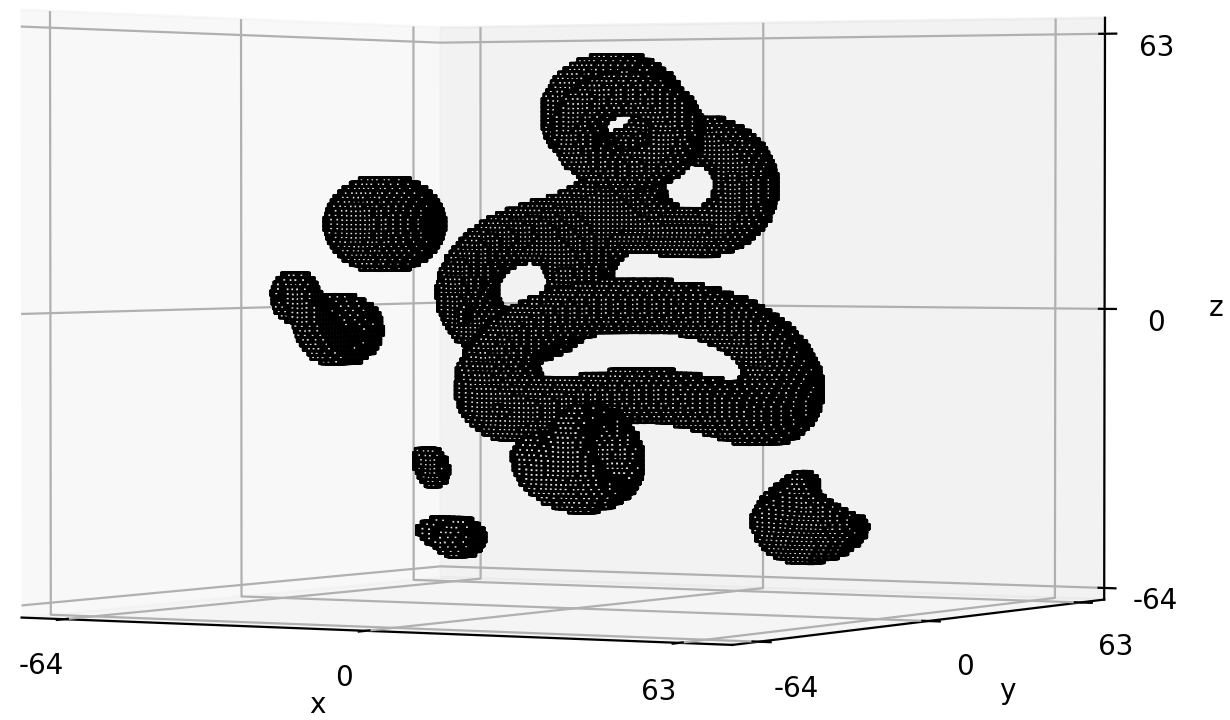}
                \end{minipage}
               
                \begin{minipage}[b]{0.45\textwidth}
                    \centering
                    \includegraphics[width=0.95\textwidth, height=0.7\textwidth]{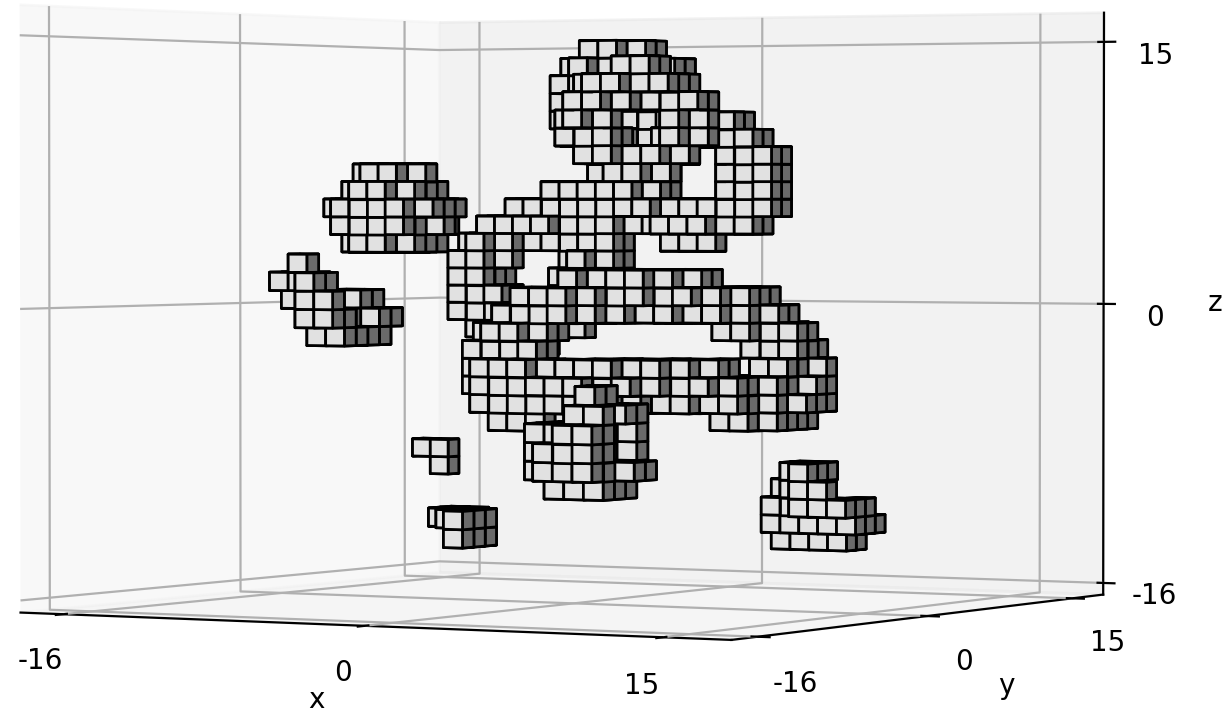}
                \end{minipage}
                                
                \caption{A 3D slice taken along the $w$-axis of a $128^4$ 4D sample before (top) and after (bottom) downsampling to $32^4$.}
                \Description{A before-and-after comparison of downsampling.}
        		\label{fig:downsampling}
            \end{figure}

            The data were downsampled from $128^4$ to $32^4$ by taking the voxels at every 4\textsuperscript{th} coordinate along each axis; the resulting images were 256 times smaller. Figure~\ref{fig:downsampling} gives an indication of how coarse-looking a 3D slice of a 4D sample becomes after the application of this degree of downsampling. Labels were not modified, however, the impact of downsampling was investigated using GUDHI. This was performed in parallel on a HPC Grid in 50-sample batches, over 640 nodes. An average of 0.66 hours was required to analyse each batch, and the entire process utilised approximately 420.14 HPC hours. Every sample remained as single-component image ($\beta_0 = 1$) after downsampling, however, the number of samples with a structure that was consistent with their label for $\beta_1$, $\beta_2$, and $\beta_3$ was markedly affected. Of the 32000 samples, only 16065 (50.2\%), 14154 (44.23\%), and 4121 (12.88\%) samples retained a structure that matched their label for each respective Betti number (Table~\ref{tab:gudhi_analysis_DS}). Only 2175 (6.8\%) samples retained a structure that was completely consistent with their label; this result is recorded in the `complete match' column.

            \begin{table}[ht!]
                \caption{GUDHI analysis of downsampled data}
                \label{tab:gudhi_analysis_DS}
                \centering
                \begin{tabular}{cccccc}
                    \toprule
                    \multirow{2}{*}{$\beta_0$} & \multirow{2}{*}{$\beta_1$} & \multirow{2}{*}{$\beta_2$} & \multirow{2}{*}{$\beta_3$} & Combined & Complete \\
                    &&&& accuracy & match \\
                    \midrule
                    100.0 & 50.2 & 44.23 & 12.88 & 51.83 & 6.8 \\ 
                    
                    \bottomrule 
                \end{tabular}
            \end{table}
                 
\begin{figure*}
                \centering
                \includegraphics[height=0.32\textwidth, trim={1cm 1cm 0 0},clip]{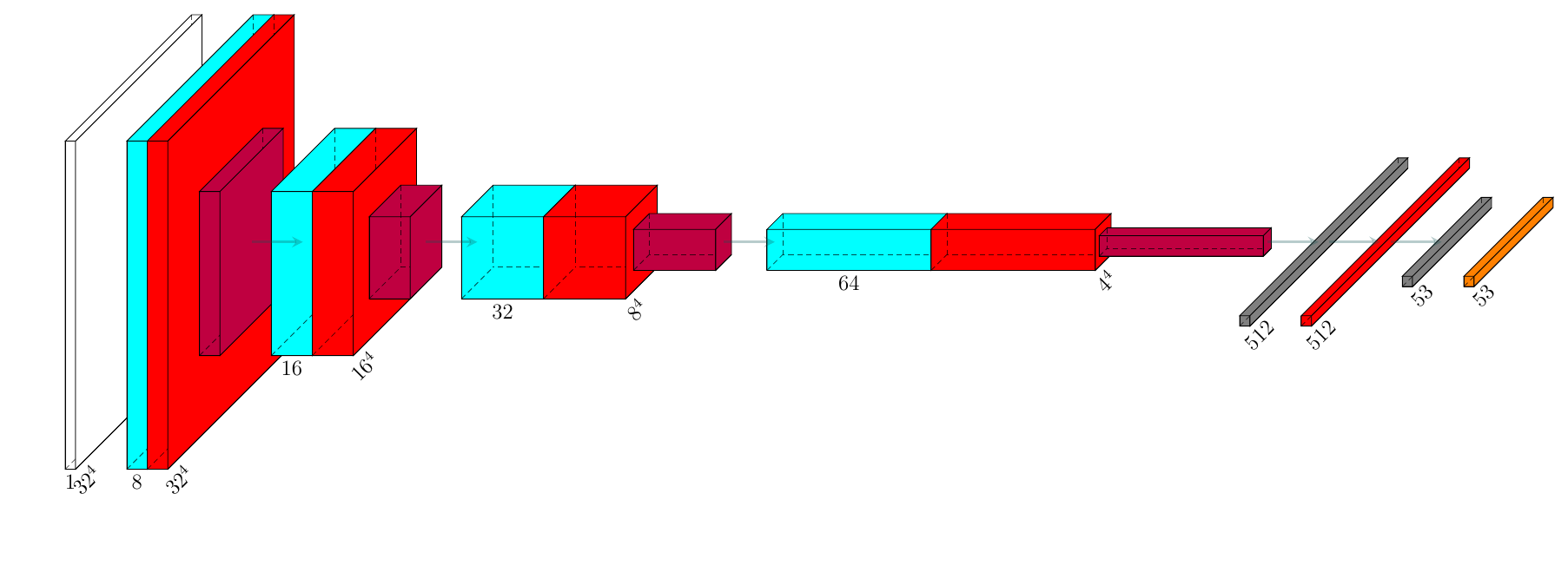}
                \caption{A visualisation of the 4D CNN that was trained with the downscaled 4D data. The input (white) passes through four iterations of a module comprising of a convolution layer (cyan), a ReLU operation (red), and a max-pooling layer (purple), before passing through two fully connected layers (grey) that are separated by a ReLU operation. The result was output to a sparsely coded vector (orange). The number of channels are denoted by the horizontal numbers, and the spatial dimensions of the inputs and feature maps are denoted by the slanted numbers. }
                \Description{A visualisation of the 4D CNN that was trained with the downscaled 4D data.}
                \label{fig:4D_CNN_downscaled}
            \end{figure*}

            The downsampled data was then used to train a 4D CNN, which was implemented with PyTorch, and had a similar architecture to the 2D and 3D CNNs that were used by~\citet{Pau19}; these consisted of several convolution layers, followed by a max-pooling layer, and finishing with a sequence of fully connected linear layers. Our 4D model is depicted in Figure~\ref{fig:4D_CNN_downscaled} and began with four iterations of a module consisting of:
            a 4D convolution layer, followed by a ReLU activation function, and then a 4D max-pooling layer.
            The convolutional layers output 8, 16, 32, and 64 channels, respectively, with 2 units of padding and a $5^4$ kernel. The pooling kernel was $2^4$ units.
            After the fourth convolution module, the result was flattened, and then passed through two fully connected layers that were separated by a ReLU operation.
            Finally, the result was output to a sparsely coded vector by reserving one output neuron for each possible value of $\beta_n$. Based on the design of our dataset, we accommodated for the values 0 and 1 for $\beta_0$, and accommodated for the values 0 to 16 for $\beta_1$, $\beta_2$, and~$\beta_3$. 

            Full-scale deep learning experiments were performed on the above-mentioned NVIDIA DGX Station using a multi-GPU arrangement. A high-bandwidth connection between the GPUs, and between the GPUs and CPU, via NVLink, made it possible to consider these devices as a single, larger, computing element, which was able to accommodate the CNN and allow for a 192-sample batch size. 
            
            The dataset was randomly divided into 90\% training, 5\% validation, and 5\% test sets at the beginning of each experiment. 
            For each epoch, the samples were rotated in random multiples of 90 degrees through a randomly selected coordinate plane as they were fed into the Pytorch dataloader; this offered twelve possible variations on each sample. The Cross Entropy loss function was appropriately set up to handle the four separate outputs for $\beta_0, \ \beta_1,\ \beta_2$, and $\beta_3$. The Adam optimiser was initialised with a learning rate of 0.001, and a scheduler was employed to reduce the learning rate by a factor of 10 at epochs 160 and 190 over a 200 epoch training schedule.
            
            Table~\ref{tab:CNN_results_downsampling} presents the test set accuracy average $\mu$ and standard deviation $\sigma$ that were achieved in five repeats of the experiment. Each experiment required just over 4 days (approximately 98 hours) to complete, and utilised its own randomly selected training set, validation set, and test set in the proportions set out above. An average combined test set accuracy of 82.41\% was achieved. 
            
            \begin{table}[ht!]
                \caption{Summary of CNN downsampling test set accuracy}
                \label{tab:CNN_results_downsampling}
                \centering
                \begin{tabular}{lccccc}
                    \toprule
                    \multirow{2}{*}{Run} & \multirow{2}{*}{$\beta_0$} & \multirow{2}{*}{$\beta_1$} & \multirow{2}{*}{$\beta_2$} & \multirow{2}{*}{$\beta_3$} & Combined \\
                    &&&&& accuracy \\
                    \midrule
                    1 & 100.0 & 76.97 & 61.92 & 78.18 & 79.27 \\ 
                    2 & 100.0 & 83.85 & 74.19 & 83.04 & 85.27 \\ 
                    3 & 100.0 & 72.45 & 60.94 & 77.6 & 77.75 \\ 
                    4 & 100.0 & 87.62 & 68.52 & 87.09 & 85.81 \\ 
                    5 & 100.0 & 85.13 & 65.34 & 85.3 & 83.94 \\ 
                    \midrule
                    
                    $\mu$ & 100.0 & 81.2 & 66.18 & 82.25 & 82.41 \\ 
                    $\sigma$ & 0.0 & 5.62 & 4.82 & 3.78 & 3.28 \\ 
                    \bottomrule
                \end{tabular}
            \end{table} 
     
        \subsection{Average-pooling approach}\label{subsec:average-pooling_approach}
\begin{figure*}[ht!]
                \centering
                \begin{tabular}{c|c}
                    Downsampled & Average-pooled \\
                    \hline
                    & \\

                    \includegraphics[trim=0.3cm 0.3cm 0.3cm 0.2cm,clip,width=0.21\textwidth]{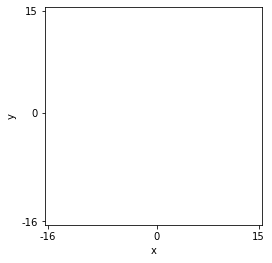}
                    \includegraphics[trim=0.3cm 0.3cm 0.3cm 0.2cm,clip,width=0.21\textwidth]{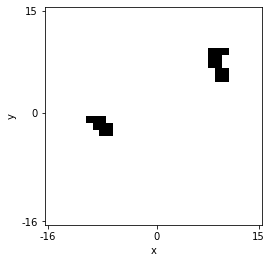}
                    &
                    \includegraphics[trim=0.3cm 0.3cm 0.3cm 0.2cm,clip,width=0.21\textwidth]{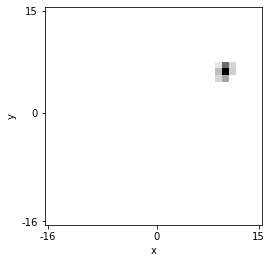}
                    \includegraphics[trim=0.3cm 0.3cm 0.3cm 0.2cm,clip,width=0.21\textwidth]{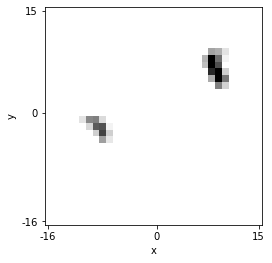}
                    \\

                    \includegraphics[trim=0.3cm 0.3cm 0.3cm 0.2cm,clip,width=0.21\textwidth]{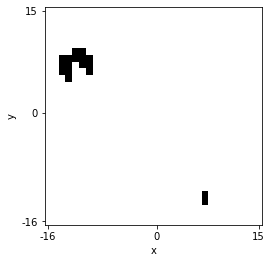}
                    \includegraphics[trim=0.3cm 0.3cm 0.3cm 0.2cm,clip,width=0.21\textwidth]{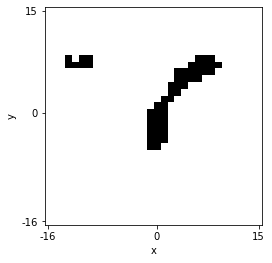}
                    &
                    \includegraphics[trim=0.3cm 0.3cm 0.3cm 0.2cm,clip,width=0.21\textwidth]{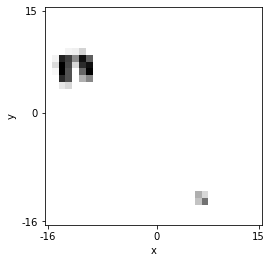}
                    \includegraphics[trim=0.3cm 0.3cm 0.3cm 0.2cm,clip,width=0.21\textwidth]{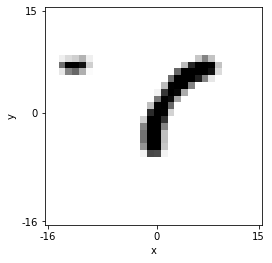}
                    \\

                    \includegraphics[trim=0.3cm 0.3cm 0.3cm 0.2cm,clip,width=0.21\textwidth]{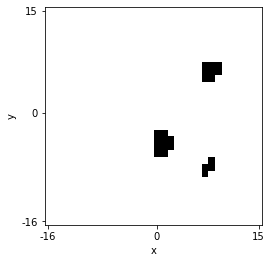}
                    \includegraphics[trim=0.3cm 0.3cm 0.3cm 0.2cm,clip,width=0.21\textwidth]{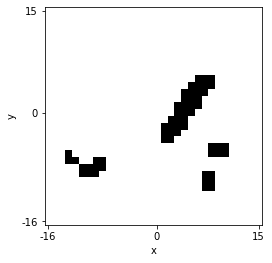}
                    &
                    \includegraphics[trim=0.3cm 0.3cm 0.3cm 0.2cm,clip,width=0.21\textwidth]{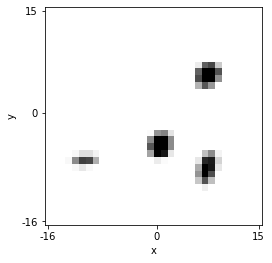}
                    \includegraphics[trim=0.3cm 0.3cm 0.3cm 0.2cm,clip,width=0.21\textwidth]{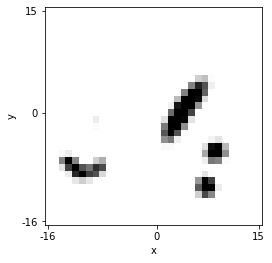}
                    \\

                    \includegraphics[trim=0.3cm 0.3cm 0.3cm 0.2cm,clip,width=0.21\textwidth]{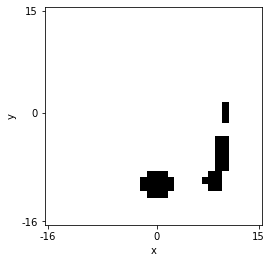}
                    \includegraphics[trim=0.3cm 0.3cm 0.3cm 0.2cm,clip,width=0.21\textwidth]{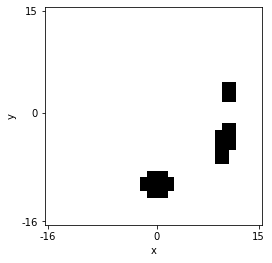}
                    &
                    \includegraphics[trim=0.3cm 0.3cm 0.3cm 0.2cm,clip,width=0.21\textwidth]{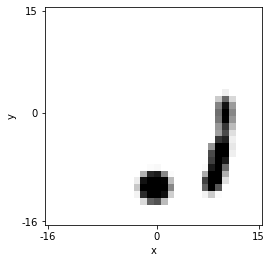}
                    \includegraphics[trim=0.3cm 0.3cm 0.3cm 0.2cm,clip,width=0.21\textwidth]{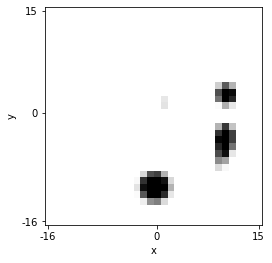}
                    \\

                    \includegraphics[trim=0.3cm 0.3cm 0.3cm 0.2cm,clip,width=0.21\textwidth]{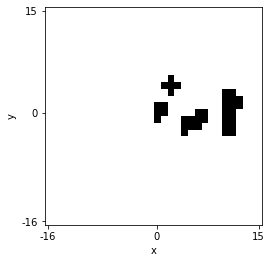}
                    \includegraphics[trim=0.3cm 0.3cm 0.3cm 0.2cm,clip,width=0.21\textwidth]{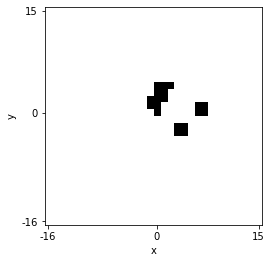}
                    &
                    \includegraphics[trim=0.3cm 0.3cm 0.3cm 0.2cm,clip,width=0.21\textwidth]{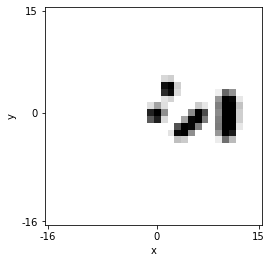}
                    \includegraphics[trim=0.3cm 0.3cm 0.3cm 0.2cm,clip,width=0.21\textwidth]{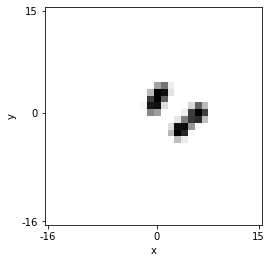}
                    \\
                \end{tabular}

                \caption{10 equally-spaced 2D slices that have been taken from the 3D slice of the 4D sample that is seen in Figure~\ref{fig:downsampling}. These slices are ordered from left to right and top to bottom. The average-pooled slices are noticeably less coarse, and, in some cases, contain more features.}
                \Description{A comparison of ten 2D slices after downsampling and average-pooling.}
        		\label{fig:visualising_avgpooled_sample}
            \end{figure*}

            Similar experiments were performed with another downscaled dataset that was produced by reducing the same $128^4$ samples to $32^4$ via 4D average-pooling; the software that was used to perform this version of downscaling was implemented with Pytorch. The labels were, again, left unaltered.
            The resulting samples were a smaller, blurry, grey-scale image of the original cube.
            Under 1\% of the average-pooled samples retained a structure that was completely consistent with their label (Table~\ref{tab:gudhi_analysis_AP}).   
            Figure~\ref{fig:visualising_avgpooled_sample} depicts a sequence of 2D slices, each taken from the $32^4$ cubes that were produced by downsampling and average-pooling the same 4D sample that was the subject of Figure~\ref{fig:downsampling}. 
            For comparison, Figure~\ref{fig:visualising_avgpooled_sample} shows equally-spaced 2D slices (taken from bottom to top) of the same downsampled $32^3$ 3D slice that is seen in the left of Figure~\ref{fig:downsampling}, and compares these slices with their corresponding average-pooled slice.
            The average-pooled slices appear to be less coarse and richer in features, versus their corresponding downsampled slice. For example, the first downsampled slice is empty, however, the average-pooled slice contains evidence of a cavity; the fifth, sixth, and eighth slices demonstrate something similar. The seventh and tenth downsampled slices contain pairs of features that are actually part of the same cavity, which we deduce by inspecting the corresponding average-pooled slice.
            This occurs because, while downsampling only collects the value (0 or 1) of the voxels at every fourth coordinate, the $4^4$ average-pooling kernel (with a 4-unit stride) takes an average of the $256$ voxel values that it sees.
            
            \begin{table*}[ht!]
                
                \caption[GUDHI analysis of average-pooled data]{GUDHI analysis of average-pooled 4D data.}
                \label{tab:gudhi_analysis_AP}
                \centering
                \begin{tabular}{cccccc}
                    \hline
                    $\beta_0$ & $\beta_1$ & $\beta_2$ & $\beta_3$ & Combined accuracy & Complete match\\
                    \hline
                    99.56 & 30.26 & 25.02 & 2.02 & 39.21 & 0.8 \\ 
                    
                    \hline
                \end{tabular}
                
            \end{table*}

            The dataset division percentages, CNN architecture, and scheduler that were detailed in Section~\ref{subsec:downsampling_approach} were also used for these experiments, however in this case, the Adam optimiser was initialised with a learning rate of 0.0001 in response to inconsistent results that were observed during several abbreviated preliminary test runs with a learning rate of 0.001. Despite the smaller learning rate, the scheduler performed the same 200 epochs. The final results of these experiments are presented in Table~\ref{tab:CNN_results_avgpool}, along with some statistics. An average combined accuracy of approximately 78.52\% was observed in these experiments. 

            \begin{table}[ht!]
                \caption{Summary of CNN average-pooling test set accuracy}
                \label{tab:CNN_results_avgpool}
                \centering
                \begin{tabular}{lccccc}
                    \toprule
                    \multirow{2}{*}{Run} & \multirow{2}{*}{$\beta_0$} & \multirow{2}{*}{$\beta_1$} & \multirow{2}{*}{$\beta_2$} & \multirow{2}{*}{$\beta_3$} & Combined \\
                    &&&&& accuracy \\
                    \midrule
                    1 & 100.0 & 61.69 & 61.57 & 82.87 & 76.53 \\ 
                    2 & 100.0 & 58.85 & 59.9 & 79.34 & 74.52 \\ 
                    3 & 100.0 & 67.59 & 67.53 & 85.3 & 80.11 \\ 
                    4 & 100.0 & 70.72 & 72.8 & 88.48 & 83.0 \\ 
                    5 & 100.0 & 66.2 & 63.89 & 83.68 & 78.44 \\ 
                    \midrule
                    
                    $\mu$ & 100.0 & 65.01 & 65.14 & 83.94 & 78.52 \\ 
                    $\sigma$ & 0.0 & 4.23 & 4.61 & 3.0 & 2.92 \\ 
                    \bottomrule
                \end{tabular}
            \end{table} 

        \subsection{Cavity-focused approach}\label{subsec:cavity-focused_approach}
            For humans, it may be more instinctive to approach the task of visually detecting holes in 2D and 3D samples by firstly identifying each cavity, and then deducing the holes that are present from this information; this overcomes the need to identify abstract features such as holes. For example, if a donut-shaped cavity is identified within a 3D sample, then this would imply the existence of one 1D hole and one 2D hole.
            
            A third set of experiments were performed using this cavity-focused approach, and employed the same models and data that were described in Sections~\ref{subsec:downsampling_approach} and~\ref{subsec:average-pooling_approach}. For these experiments, a sample's label encoded the number of times that each manifold had been removed from the original 4D cube by using a vector that was ordered 
            $[B^4, S^1 \times B^3, S^2 \times B^2, S^1 \times S^1 \times B^2]$; 
            the model's output layer was also modified to accommodate this.
            Combining the model's output with the approach that was used to derive Equation~\ref{eqn:sample_betti_numbers} makes it possible to then estimate the Betti numbers of a sample.
            
            For the cavity-focused downsampling experiment, the Adam optimiser was initialised with a learning rate of 0.001, and a scheduler was employed to reduce the learning rate by a factor of 10 at epochs 80 and 95 over a 100 epoch training schedule.
            Table~\ref{tab:CNN_results_cavityfocused_DS_cavities} presents the test set results that were achieved in five repeats of this experiment, and Table~\ref{tab:CNN_results_cavityfocused_DS_betti_numbers} presents the same results after converting the CNN's outputs to Betti numbers; this permits a direct comparison with the results that are presented in Tables~\ref{tab:CNN_results_downsampling} and \ref{tab:CNN_results_avgpool}.

            \begin{table}[ht!]
                \caption{Summary of CNN cavity-focused downsampling test set accuracy - cavities}
                \label{tab:CNN_results_cavityfocused_DS_cavities}
                \centering
                \begin{tabular}{lccccc}
                    \toprule
                    \multirow{2}{*}{Run} & \multirow{2}{*}{$B^4$} & \multirow{2}{*}{$S^1 \times B^3$} & \multirow{2}{*}{$S^2 \times B^2$} & \multirow{2}{*}{$S^1 \times S^1 \times B^2$} & Combined \\
                    &&&&& accuracy\\
                    \midrule
                    1 & 96.7 & 90.74 & 90.05 & 93.0 & 92.62 \\ 
                    2 & 96.99 & 93.75 & 93.17 & 94.56 & 94.62 \\ 
                    3 & 93.98 & 89.53 & 85.24 & 91.44 & 90.05 \\ 
                    4 & 95.43 & 92.59 & 89.18 & 89.58 & 91.7 \\ 
                    5 & 97.11 & 89.24 & 88.08 & 92.19 & 91.65 \\ 
                    \hline
                    
                    $\mu$ & 96.04 & 91.17 & 89.14 & 92.15 & 92.13 \\ 
                    $\sigma$ & 1.19 & 1.75 & 2.58 & 1.65 & 1.5 \\
                    \bottomrule
                \end{tabular}
            \end{table} 

            The cavity-focused downsampling CNN achieved greater than 90\% accuracy when detecting cavities (Table~\ref{tab:CNN_results_cavityfocused_DS_cavities}), however, this result does not directly transfer to the network's ability to estimate Betti numbers (Table~\ref{tab:CNN_results_cavityfocused_DS_betti_numbers}). 
            In contrast to the experiments discussed in Sections~\ref{subsec:downsampling_approach} and~\ref{subsec:average-pooling_approach}, estimating Betti numbers with a cavity-focused CNN introduces a constraint on the combination of Betti numbers that are possible because they are derived deterministically from the cavities that are detected. Consequently, incorrectly detecting one cavity can result in an incorrect estimation of more than one Betti number. For example, missing a ball-shaped cavity will only affect an estimate of $\beta_3$, however, incorrectly detecting a cavity that is formed by $S^1 \times S^1 \times B^2$ will affect the estimates of $\beta_1$, $\beta_2$ and $\beta_3$.
            The results that are listed in Table~\ref{tab:CNN_results_cavityfocused_DS_betti_numbers} show that this CNN achieved slightly better, but comparable, results to those that are presented in Tables~\ref{tab:CNN_results_downsampling} and \ref{tab:CNN_results_avgpool}.

            \begin{table}[ht!]
                \caption{Summary of CNN cavity-focused downsampling test set accuracy - Betti numbers}
                \label{tab:CNN_results_cavityfocused_DS_betti_numbers}
                \centering
                \begin{tabular}{lccccc}
                    \toprule
                    \multirow{2}{*}{Run} & \multirow{2}{*}{$\beta_0$} & \multirow{2}{*}{$\beta_1$} & \multirow{2}{*}{$\beta_2$} & \multirow{2}{*}{$\beta_3$} & Combined \\
                    &&&&& accuracy \\
                    \midrule
                    1 & 100.0 & 85.07 & 84.49 & 75.98 & 86.39 \\ 
                    2 & 100.0 & 89.06 & 88.83 & 82.99 & 90.22 \\ 
                    3 & 100.0 & 78.76 & 82.35 & 70.95 & 83.02 \\ 
                    4 & 100.0 & 82.12 & 83.45 & 75.69 & 85.32 \\ 
                    5 & 100.0 & 82.99 & 82.64 & 75.12 & 85.19 \\ 
                    \hline
                    
                    $\mu$ & 100.0 & 83.6 & 84.35 & 76.15 & 86.02 \\ 
                    $\sigma$ & 0.0 & 3.41 & 2.36 & 3.88 & 2.37 \\
                    \bottomrule
                \end{tabular}
            \end{table} 

            For the cavity-focused average-pooling experiment, the Adam optimiser was initialised with a learning rate of 0.0001, and a scheduler was employed to reduce the learning rate by a factor of 10 at epochs 160 and 190 over a 200 epoch training schedule; these are the same hyperparameters that were used in the experiment that was described in Section~\ref{subsec:average-pooling_approach}. 
            Several abbreviated preliminary test runs were performed using a 100-epoch training schedule, a 150-epoch training schedule, and a learning rate of 0.001, however, these settings resulted in model underfitting and erratic training.
            Tables~\ref{tab:CNN_results_cavityfocused_AP_cavities} and~\ref{tab:CNN_results_cavityfocused_AP_betti_numbers} present the test set results that were achieved in five repeats of this experiment.
            
            \begin{table}[ht!]
                \caption{Summary of CNN cavity-focused average-pooling test set accuracy - cavities}
                \label{tab:CNN_results_cavityfocused_AP_cavities}
                \centering
                \begin{tabular}{lccccc}
                    \toprule
                    \multirow{2}{*}{Run} & \multirow{2}{*}{$B^4$} & \multirow{2}{*}{$S^1 \times B^3$} & \multirow{2}{*}{$S^2 \times B^2$} & \multirow{2}{*}{$S^1 \times S^1 \times B^2$} & Combined \\
                    &&&&& accuracy\\
                    \midrule
                    1 & 96.53 & 83.62 & 79.57 & 95.37 & 88.77 \\ 
                    2 & 96.64 & 87.79 & 81.89 & 91.96 & 89.57 \\ 
                    3 & 93.75 & 87.91 & 78.88 & 95.54 & 89.02 \\ 
                    4 & 95.08 & 85.53 & 75.69 & 92.25 & 87.14 \\ 
                    5 & 95.54 & 87.96 & 79.11 & 92.71 & 88.83 \\ 
                    \hline
                    
                    $\mu$ & 95.51 & 86.56 & 79.03 & 93.56 & 88.67 \\ 
                    $\sigma$ & 1.06 & 1.73 & 1.98 & 1.56 & 0.81 \\ 
                    \bottomrule
                \end{tabular}
            \end{table}
            
            In comparison to the results that are provided in Table~\ref{tab:CNN_results_avgpool}, the cavity-focused average-pooling approach resulted in models that were more accurate overall (Table~\ref{tab:CNN_results_cavityfocused_AP_betti_numbers}), although, they were less accurate when estimating $\beta_3$. The cavity-focused average-pooling CNN achieved an average combined accuracy of 88.67\% when detecting cavities (Table~\ref{tab:CNN_results_cavityfocused_AP_cavities}), which was lower versus the performance of the models that were trained using the cavity-focused downsampling approach, despite training for twice as many epochs; the model appeared to have difficulty with identifying the number of $S^2 \times B^2$-based cavities.
            An average combined accuracy of 81.86\% was achieved when estimating Betti numbers. The lower accuracies with which $\beta_1$ and $\beta_3$ were estimated were consistent with the model's difficulty with $S^2 \times B^2$.
            
            \begin{table}[ht!]
                \caption{Summary of CNN cavity-focused average-pooling test set accuracy - Betti numbers}
                \label{tab:CNN_results_cavityfocused_AP_betti_numbers}
                \centering
                \begin{tabular}{lccccc}
                    \toprule
                    \multirow{2}{*}{Run} & \multirow{2}{*}{$\beta_0$} & \multirow{2}{*}{$\beta_1$} & \multirow{2}{*}{$\beta_2$} & \multirow{2}{*}{$\beta_3$} & Combined \\
                    &&&&& accuracy \\
                    \midrule
                    1 & 100.0 & 77.2 & 80.03 & 70.31 & 81.89 \\ 
                    2 & 100.0 & 78.12 & 80.9 & 72.92 & 82.99 \\ 
                    3 & 100.0 & 77.31 & 84.43 & 69.5 & 82.81 \\ 
                    4 & 100.0 & 72.05 & 79.63 & 66.38 & 79.51 \\ 
                    5 & 100.0 & 76.79 & 81.77 & 69.85 & 82.1 \\ 
                    \hline
                    
                    $\mu$ & 100.0 & 76.3 & 81.35 & 69.79 & 81.86 \\ 
                    $\sigma$ & 0.0 & 2.17 & 1.71 & 2.09 & 1.24 \\ 
                    \bottomrule
                \end{tabular}
            \end{table}

        \subsection{Efficacy of CNN-based Betti Number Estimation}\label{subsec:efficacy}
            In order to better understand how our different approaches to Betti number estimation compared to one another, 2000 additional $128^4$ samples were generated and then downscaled via the downsampling and average-pooling approaches that were described in Sections~\ref{subsec:downsampling_approach} and \ref{subsec:average-pooling_approach}; these data were therefore new to all the trained models. As expected, an analysis of these datasets, using both GUDHI and the trained CNNs, produced similar results to those presented in Tables~\ref{tab:gudhi_analysis_DS}, \ref{tab:CNN_results_downsampling}, \ref{tab:CNN_results_avgpool}, \ref{tab:CNN_results_cavityfocused_DS_betti_numbers}, and~\ref{tab:CNN_results_cavityfocused_AP_betti_numbers}.

            The architecture of the CNNs and many of the training hyperparameters were left unchanged across all of the training experiments. While benchmarking the proposed downscaling and training options was not the primary aim of this work, it was observed that some of the experiments resulted in CNNs that exhibited particular strengths.
            Despite the significant inconsistency between the topological structure of these downscaled images and their labels (only 6.35\% of the new samples exhibited a complete label match after downsampling, as demonstrated by GUDHI), the CNNs were still able to achieve a complete match for over 50\% of the samples and more accurately estimated the respective Betti numbers of each sample versus the results that were produced via persistent homology (Table~\ref{tab:comparison_of_performance}).
    
            We also observed that average-pooling produced a slightly better $\beta_3$ estimation, which may have been a consequence of the richer-looking data (as demonstrated in Figure~\ref{fig:visualising_avgpooled_sample}), although, this CNN also performed worse when estimating~$\beta_1$.
            
            Both CNNs that were trained with a cavity focus demonstrated an improvement in complete-match rates.
            The cavity-focused downsampling CNN performed significantly better than the alternatives, achieving a combined accuracy of over 90\% and a complete-match rate of over 80\%.
            

            \begin{table*}[ht!]
                
                \caption{Comparing CNN performance when using downsampling (DS), average-pooling (AP), and cavity-focused (CF) approaches. The `Time' column refers to the time required to analyse the dataset.}
                \label{tab:comparison_of_performance}
                \centering
                \renewcommand{\arraystretch}{1.2}
                \begin{tabular}{lccccccc}
                    \toprule
                    & \multirow{2}{*}{$\beta_0$} & \multirow{2}{*}{$\beta_1$} & \multirow{2}{*}{$\beta_2$} & \multirow{2}{*}{$\beta_3$} & Combined  & Complete & Time\\
                    &&&&& accuracy & match & (hours)\\
                    \midrule
                    GUDHI - DS & 100.0 & 50.4 & 44.3 & 12.15 & 51.73 & 6.35 & 15.5\\ 
                    \midrule
                    GUDHI - AP & 99.6 & 28.65 & 24.65 & 2 & 38.73 & 1.05 & 12 \\ 
                    \midrule
                    CNN - DS & 100.0 & 85.25 & 67.85 & 86.95 & 85.01 & 53.5 & 0.08 \\ 
                    \midrule
                    CNN - AP & 100.0 & 70.65 & 72.65 & 88.2 & 82.88 & 50.55 & 0.07 \\ 
                    \midrule
                    CNN - CFDS & 100.0 & 88.85 & 88.85 & 82.45 & 90.04 & 80.4 & 0.1\\ 
                    \midrule
                    CNN - CFAP & 100.0 & 76.4 & 82.1 & 71.2 & 82.43 & 65.25 & 0.1\\ 
                    \bottomrule
                \end{tabular}
                
            \end{table*}

    \section{A study using 3D real-world data}\label{sec:a_study_using_3d}
        Due to computational demands, 4D data processing is presently only possible on data of very restricted size, even with the above mentioned NVIDIA DGX Station.
        We are not aware of any real-world datasets that comprise of 3D image-type data with labels that correlate with the Betti numbers of these data. 
        Therefore, an unlabelled 3D dataset with real-world characteristics was sought, with the aim to use them to complement the results on synthetic 4D data.
        This section details how existing 3D real-world data was processed in order to produce a dataset of labelled 3D samples, with which it was possible to demonstrate how a 3D CNN could perform in a topology estimation task corresponding to the 4D tasks which were addressed in the preceding sections when prepared with unscaled (original resolution) and downscaled data. As in the 4D case, the 3D results are compared to those obtained using persistent homology calculations.
        
        \subsection{3D dataset generation}\label{subsec:3d_data_gen}
\begin{figure}[ht!]
                \centering
                \includegraphics[width=0.4\textwidth]{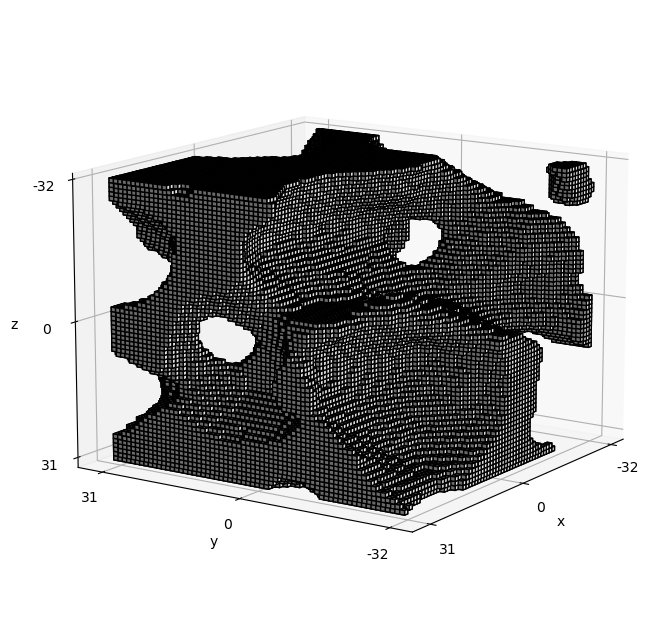}
                \includegraphics[width=0.4\textwidth]{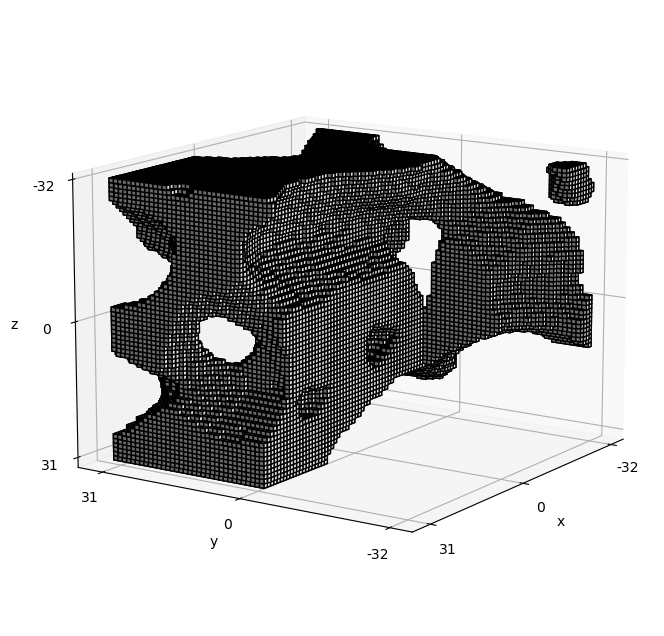}
                \caption{An example of a 3D sample that was cut out of a larger real-world sample (left image). This sample contains two components, two 1D holes, and three 2D holes; two of the 2D holes are shown in the cutaway (right image). That is, $\beta_0$ and $\beta_1$ are 2, and $\beta_2$ is 3. }
                \Description{An example of a real-world 3D sample.}
                \label{fig:3d_S10}
            \end{figure}

            The real-world data that was employed in these experiments was collected as part of the work by~\citet{Fie20} and consists of a CT scan of a manufactured metallic syntactic foam. The dimensions of this 3D grey-scale image are $684 \times 744 \times 887$, with pixel values ranging from 0 (black) to 255 (white). 
            A collection of 32000 $64^3$ cubes were cut out from randomly selected locations of this image. Each cube was converted into a black-and-white image by using a grey threshold that was randomly chosen between the values of 60 and 80. 
            This range was based on software testing observations, which uncovered that a threshold within this range retained most of the original structure of samples that were cut out from this particular CT scan. This was because the pixels that represented the foam material took on similar values; very low or high threshold values resulted in exaggerated features and, in extreme cases, primarily black images, which would have been topologically trivial.
            Standard image preprocessing morphology techniques, such as erosion and dilation, were then applied in order to remove single-pixel artefacts. It was possible to generate a Betti number label for these data by analysing them with GUDHI. In this case,
            the labels took the form $[\beta_0, \beta_1, \beta_2]$. A secondary label was also included, which encoded the Betti numbers for the inverse of each cube. The values of the Betti numbers ranged from 0 to 8.
            It is noteworthy that the samples comprising this dataset are not limited to single components; a sample with two components is shown in Figure~\ref{fig:3d_S10}.
            While it has been shown that CNNs are capable of segmenting (up to three) objects, such as spheres and multi-holed tori, in 3D point-cloud data~\citep{Pee23},
            homology is not completely discriminative.
            For example, consider one 3D cube with cavities that are formed by removing a 2-holed donut and a ball, and consider another cube with cavities that are formed by removing two donuts. Notice that both cubes would have the label [1,2,2]. Furthermore, considering the complements of the cubes does not help to distinguish between the samples in this example because both complements share the label [2,2,0]. This suggests that additional work would be needed to generate the labels that could be used for a cavity-focused training scheme.
            
            The $64^3$ samples were subsequently downscaled by applying downsampling and average-pooling in order to produce two datasets comprising of $16^3$ samples. The effect of downscaling on the structure of the samples is summarised in Table~\ref{tab:3d_gudhi}, which demonstrates that persistent homology appears to tolerate downsampling more so than average-pooling, similarly as seen in the 4D results.

            \begin{table}[ht!]
                
                \caption[GUDHI analysis of 3D data]{GUDHI analysis of the downsampled (DS) and average-pooled (AP) 3D data.}
                \label{tab:3d_gudhi}
                \centering
                \begin{tabular}{lccccc}
                    \toprule
                    & \multirow{2}{*}{$\beta_0$} & \multirow{2}{*}{$\beta_1$} & \multirow{2}{*}{$\beta_2$} & Combined  & Complete\\
                    &&&& accuracy & match\\
                    \midrule
                    GUDHI - DS & 55.34 & 53.83 & 79.03 & 62.73 & 23.72 \\
                    \midrule
                    GUDHI - AP & 29.81 & 22.56 & 70.8 & 41.06 & 5.08 \\
                    \bottomrule
                \end{tabular}
                
            \end{table}

        \subsection{3D dataset distribution}
            The explicit distribution of the Betti numbers of the acquired 3D dataset are summarised in Table~\ref{tab:3d_dataset_dist_betti}. 
            The statistics demonstrate that a majority of the samples did not contain a 2D hole; if they did, it was most likely only one or two. Similarly, it was much less likely for a sample to contain a higher number of components or 1D holes.

            \begin{table*}[ht!]
                
                \caption{The percentage of samples with a respective Betti number label. The value of the Betti numbers ranged from 0 to 8.}
                \label{tab:3d_dataset_dist_betti}
                \centering
                \begin{adjustbox}{max width=\textwidth}
                \begin{tabular}{lccccccccc}
                    \toprule 
                    $\beta_n$ & 0 & 1 & 2 & 3 & 4 & 5 & 6 & 7 & 8 \\
                    \midrule
                    0 & 0 & 27.72 & 32.55 & 21.5 & 10.95 & 4.59 & 1.78 & 0.68 & 0.23 \\
                    1 & 4.63 & 13.92 & 20.82 & 21.12 & 16.48 & 11.05 & 6.43 & 3.68 & 1.88 \\
                    2 & 77.24 & 17.92 & 3.76 & 0.79 & 0.18 & 0.064 & 0.025 & 0.016 & 0.003 \\
                    
                    \bottomrule
                \end{tabular}
                \end{adjustbox}
                
            \end{table*}

        \subsection{Experiments and results}\label{sec:experiments_and_results_3d}
            A series of deep learning experiments in the task of estimating Betti numbers were performed, the first of which employed the unscaled data and the remaining two of which employed the downsampled and average-pooled data. The experiments were performed on the above-mentioned NVIDIA DGX Station.

\begin{figure*}
                \centering
                \includegraphics[height=0.32\textwidth, trim={1cm 1cm 0 0},clip]{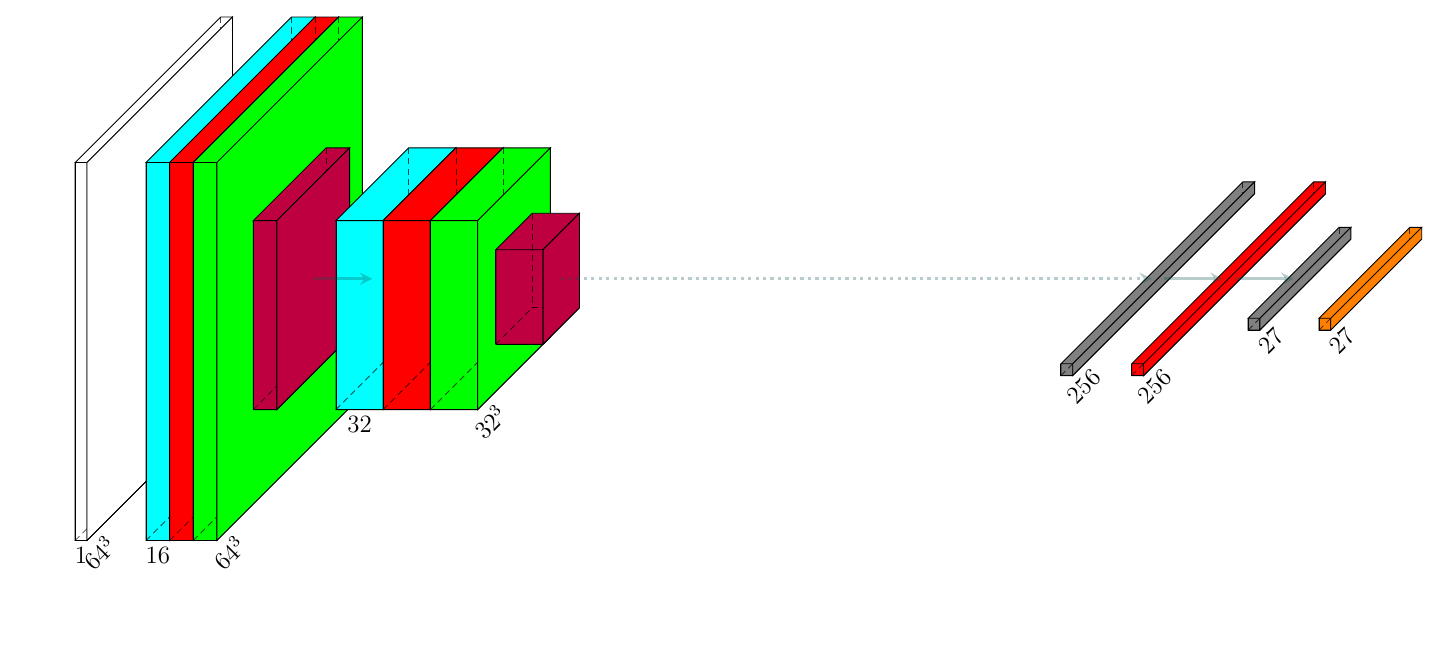}
                \caption{A visualisation of the 3D CNN that was trained with the unscaled 3D data. The input (white) passes through six iterations of a module comprising of a convolution layer (cyan), a ReLU operation (red), a batch normalisation layer (green), and a max-pooling layer (purple). Due to space constraints, only two iterations are shown here. The remaining iterations would lie along the dotted line. The result then passes through two fully connected layers (grey) that are separated by a ReLU operation. The result was output to a sparsely coded vector (orange). The number of channels are denoted by the horizontal numbers, and the spatial dimensions of the inputs and feature maps are denoted by the slanted numbers. }
                \Description{A visualisation of the 3D CNN that was trained with the unscaled 3D data.}
                \label{fig:3D_CNN_unscaled}
            \end{figure*}

            The unscaled data was used to train a basic CNN model that is depicted in Figure~\ref{fig:3D_CNN_unscaled} and began with six iterations of a module consisting of:
            a 3D convolution layer, followed by a ReLU activation function, a batch normalisation layer, and then a max-pooling layer.
            The convolutional layers output 16, 32, 64, 128, 256, and 512 channels, respectively, with 1 unit of padding and a $3^3$ kernel. The pooling kernel was $2^3$ units.
            After the sixth convolution module, the result was flattened, and then passed through two fully connected layers that were separated by a ReLU operation.
            Finally, the result was output to a sparsely coded vector that accommodated for the values 0 to 8 for each Betti number.
            The dataset was randomly divided into 90\% training, 5\% validation, and 5\% test sets at the beginning of each experiment. A 32-sample batch size was used. 
            For each epoch, the samples were rotated in random multiples of 90 degrees through a randomly selected coordinate plane as they were fed into the Pytorch dataloader. The Cross Entropy loss function was appropriately set up to handle the three separate outputs. The Adam optimiser was initialised with a learning rate of 0.0025, and a scheduler was employed to reduce the learning rate by a factor of 10 at epochs 80 and 90 over a 100 epoch training schedule; this took approximately 2.1 hours to complete.
            Table~\ref{tab:3d_CNN_results_unscaled} presents the test set accuracy average $\mu$ and standard deviation $\sigma$ that were achieved. An average combined test set accuracy of 93.36\% was achieved.

            \begin{table}[ht!]
                
                \caption{Summary of 3D CNN unscaled test set accuracy}
                \label{tab:3d_CNN_results_unscaled}
                \centering
                \begin{tabular}{lccccc}
                    \toprule
                    \multirow{2}{*}{Run} & \multirow{2}{*}{$\beta_0$} & \multirow{2}{*}{$\beta_1$} & \multirow{2}{*}{$\beta_3$} & Combined & Complete \\
                    &&&& accuracy & match \\
                    \midrule
                    1 & 96.63 & 86.19 & 97.56 & 93.46 & 82.0 \\ 
                    2 & 95.94 & 87.5 & 98.56 & 94.0 & 83.44 \\ 
                    3 & 94.94 & 88.25 & 98.06 & 93.75 & 82.88 \\ 
                    4 & 95.75 & 85.31 & 98.38 & 93.15 & 81.0 \\ 
                    5 & 95.0 & 84.63 & 97.75 & 92.46 & 79.75 \\ 
                    \midrule
                    
                    $\mu$ & 95.65 & 86.38 & 98.06 & 93.36 & 81.81 \\ 
                    $\sigma$ & 0.63 & 1.34 & 0.37 & 0.53 & 1.32 \\ 
                    \bottomrule
                \end{tabular}
                
            \end{table}

\begin{figure*}
                \centering
                \includegraphics[height=0.32\textwidth, trim={1cm 1cm 0 0},clip]{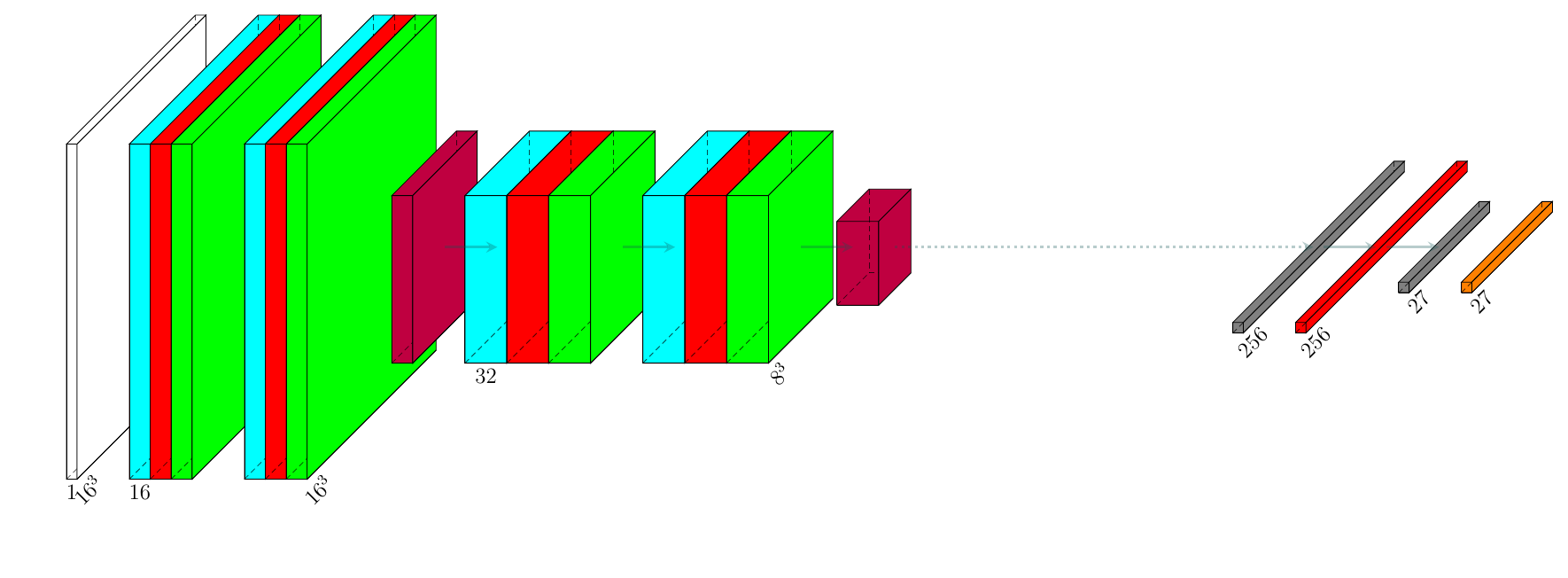}
                \caption{A visualisation of the 3D CNN that was trained with the downscaled 3D data. 
                The input (white) passes through eight iterations of a module comprising of a convolution layer (cyan), a ReLU operation (red), and a batch normalisation layer (green). A max-pooling layer (purple) was applied after the even-numbered modules. 
                Due to space constraints, only four iterations are shown here. The remaining iterations would lie along the dotted line. The result then passes through two fully connected layers (grey) that are separated by a ReLU operation. The result was output to a sparsely coded vector (orange). The number of channels are denoted by the horizontal numbers, and the spatial dimensions of the inputs and feature maps are denoted by the slanted numbers. }
                \Description{A visualisation of the 3D CNN that was trained with the downsscaled 3D data.}
                \label{fig:3D_CNN_downscaled}
            \end{figure*}

            The downscaled data were used to train a model that is depicted in Figure~\ref{fig:3D_CNN_downscaled} and began with eight iterations of a module consisting of:
            a 3D convolution layer, followed by a ReLU activation function, and then 
            a batch normalisation layer. 
            A max-pooling layer was applied after the even-numbered modules; the pooling kernel was $2^3$ units.
            The first pair of convolutional layers output 16 channels, the second pair output 32 channels, and the third and fourth pairs output 64 and 128 channels, respectively, each with 1 unit of padding and a $3^3$ kernel.
            After the eighth convolution module, the result was again flattened, passed through two fully connected layers that were separated by a ReLU operation, and then output to a sparsely coded vector.
            The dataset was randomly divided into 90\% training, 5\% validation, and 5\% test sets at the beginning of each experiment. A 32-sample batch size was used and rotations were again randomly applied to the samples as they were fed into the Pytorch dataloader. The Cross Entropy loss function was set up as previously described. The Adam optimiser was initialised with a learning rate of 0.0025, and scheduled to reduce the learning rate by a factor of 10 at epochs 80 and 90 over a 100 epoch training schedule; this took approximately 1.4 hours to complete.
            
            Tables~\ref{tab:3d_CNN_results_downsampled} and~\ref{tab:3d_CNN_results_pooled} present the test set accuracy average $\mu$ and standard deviation $\sigma$ that were achieved. An average combined test set accuracy of 61.9\% and 83.16\% was achieved for the downsampled and average-pooling experiments, respectively. The CNN performed similarly to persistent homology (Table~\ref{tab:3d_gudhi}) when trained with the downsampled data, but  
            demonstrated significantly better performance versus persistent homology when trained with the average-pooled data.

            \begin{table}[ht!]
                
                \caption{Summary of 3D CNN downsampled test set accuracy}
                \label{tab:3d_CNN_results_downsampled}
                \centering
                \begin{tabular}{lccccc}
                    \toprule
                    \multirow{2}{*}{Run} & \multirow{2}{*}{$\beta_0$} & \multirow{2}{*}{$\beta_1$} & \multirow{2}{*}{$\beta_3$} & Combined & Complete \\
                    &&&& accuracy & match \\
                    \midrule
                    1 & 54.25 & 50.56 & 80.81 & 61.88 & 21.88 \\ 
                    2 & 53.56 & 51.25 & 81.0 & 61.94 & 21.81 \\ 
                    3 & 55.5 & 50.75 & 80.44 & 62.23 & 22.31 \\ 
                    4 & 52.62 & 50.62 & 79.5 & 60.92 & 21.38 \\ 
                    5 & 55.56 & 50.94 & 81.06 & 62.52 & 24.19 \\ 
                    \midrule
                    
                    $\mu$ & 54.3 & 50.82 & 80.56 & 61.9 & 22.31 \\ 
                    $\sigma$ & 1.13 & 0.25 & 0.57 & 0.54 & 0.98 \\
                    \bottomrule
                \end{tabular}
                
            \end{table}

            \begin{table}[ht!]
                
                \caption{Summary of 3D CNN average-pooled test set accuracy}
                \label{tab:3d_CNN_results_pooled}
                \centering
                \begin{tabular}{lccccc}
                    \toprule
                    \multirow{2}{*}{Run} & \multirow{2}{*}{$\beta_0$} & \multirow{2}{*}{$\beta_1$} & \multirow{2}{*}{$\beta_3$} & Combined & Complete \\
                    &&&& accuracy & match \\
                    \midrule
                    1 & 87.75 & 73.0 & 89.88 & 83.54 & 58.13 \\ 
                    2 & 86.38 & 70.62 & 89.69 & 82.23 & 55.62 \\ 
                    3 & 88.88 & 73.06 & 88.25 & 83.4 & 58.63 \\ 
                    4 & 87.31 & 72.25 & 89.88 & 83.15 & 57.5 \\ 
                    5 & 88.62 & 72.06 & 89.75 & 83.48 & 57.75 \\ 
                    \midrule
                    
                    $\mu$ & 87.79 & 72.2 & 89.49 & 83.16 & 57.52 \\ 
                    $\sigma$ & 0.91 & 0.88 & 0.62 & 0.48 & 1.02 \\ 
                    \bottomrule
                \end{tabular}
                
            \end{table}

    \section{Discussion}\label{sec:discussion}
        Provided the availability of appropriate computing resources, and
        if prior knowledge or preliminary analysis of data identifies that features of interest, such as the holes, are large enough or captured in a high enough resolution to tolerate downscaling, then persistent homology may still be a suitable option to determine the topology by calculating the Betti numbers; for example, this may be true when analysing materials that are manufactured under known conditions or to specification.

        If synthetic data, which sufficiently models 4D real-world data, is acquired, then a computer vision approach using CNNs may be suitable to estimate the Betti numbers of the data; the 4D data acquisition may follow similar efforts to those in the 3D data generation context~\citep{Gao22, Bis22, Bis24, Ken22}.
        The application of downscaling may prove to be useful in cases where it may be less critical to determine the exact topology of data, such as in some areas of material science, or
        where it is already appropriate to consider the results of persistent homology less explicitly, for example, via persistence images, as demonstrated in the analysis of data derived from dynamical systems~\citep{Ada17}.
        
        The 4D results of our work demonstrate that it is possible to apply downscaling methods prior to employing CNNs to estimate the Betti numbers of 4D image-type manifold data on which it may not be possible to directly apply existing methods due to large data size, or where downscaling is too disruptive for traditional persistent homology algorithms, such as GUDHI; our samples were reduced by a factor of 256.
        The representative comparison with persistent homology software
        demonstrates that CNNs appear to be more robust to the homological changes resulting from downscaling.
        This conclusion is supported by the results that were observed in 3D experiments, which demonstrated that a CNN could still perform well when trained with average-pooled real-world derived data. The 3D experiments also showed that CNNs can estimate the Betti numbers of multi-component samples.

        In separate 4D experiments, a cavity-focused training approach was used to compare with the results that were previously presented in the workshop paper by~\citet{Han23}. 
        A cavity-focused 4D CNN was implemented by adjusting the output layer (as described in Section~\ref{subsec:cavity-focused_approach}) and then trained under the same training schedule as those that were proposed in the workshop paper; note that it was not necessary to downscale the $32^4$-sized data of~\citet{Han23}
        in these experiments. The resulting cavity-focused CNN demonstrated a similar test set average accuracy to those presented in the workshop paper when estimating Betti numbers. The two approaches were then compared using a new 2000-sample dataset of $32^4$ samples (Section~\ref{subsec:efficacy}). The original CNN achieved an average accuracy of 95.89\% and a complete match of 85.3\%. The cavity-focused CNN performed only marginally better on these data, achieving 96.41\% and 88.85\%, respectively. This suggests that performance differences between a cavity-focused approach and a corresponding approach with Betti number targets only become apparent when the complexity (potential number of cavities and size) of the data and task is increased, as in the 4D experiments that were presented in this present work.
                
        The latter new results of our present study offer some general insight into how to train a CNN for Betti number estimation,
        which did not become apparent until increasing the complexity of the data. 
        It appears that using a cavity-focused approach results in a better performing CNN, which reveals an interesting parallel to how humans may naturally approach the same task in 2D or 3D, that is, by visually detecting cavities associated with manifolds of certain topology types as boundaries, instead of directly estimating the Betti numbers of a sample.
        The cavity-focused CNN achieved an average combined accuracy of over 90\% and a complete-match accuracy of just over 80\%, despite the significant discrepancies between the structure of downscaled training data and their original labels.
        Downsampling also granted the use of persistent homology software, however, only about 51\% and 6\% accuracy were achieved for the same metrics.

        The test set accuracies that were achieved in the 4D downsampling and average-pooling experiments were comparable to one another.
        While a noticeable difference in CNN performance was observed when downscaling was applied in the 3D experiments, neither downsampling nor average-pooling resulted in a CNN that performed worse than persistent homology.  
        Overall, this would suggest that both downsampling options could be useful with some fine-tuning.
        However, the 3D results would support the recommendation that applying average-pooling would be an appropriate first choice in cases where downscaling would be an option.
        A cavity-focused approach would be particularly useful in cases where it would be possible to acquire labelled synthetic data.

        \subsection{Limitations}\label{subsec:limitations}

            The 4D results and insights that have been presented in this work extend the approach that was pioneered by~\citet{Pau19} in the 2D and 3D setting. 
            Collectively, these results have, primarily, been restricted to synthetic, single-component samples. 
            More complicated features, such as multiple-components, links, and connected sums, or the use of topology-preserving deformations to vary the geometric appearance of cavities, are only just beginning to be considered in this line of research~\citep{Han24}; it is likely that real-world 4D data would possess these features, similarly as in the 3D data that was described in Section~\ref{sec:a_study_using_3d}.
            Addressing this would be essential in order to apply the CNN approach more generally to the task of estimating homology. 

            Although we have demonstrated that it is possible to analyse large 4D samples with CNNs where using existing options may not be feasible, it is apparent from our work that the process of generating synthetic data with which to train a CNN comes at a significant resource and time cost that may not be accessible to everyone. Choosing how to generate data that models real-world data may also be a necessary preliminary step, which brings with it a new set of challenges such as quantifying the differences between geometric shapes or textures~\citep{Tur14}. 
            
            In our case, we faced the complication of finding a balance between the size of the data that we considered and the size of the CNN that we used. The decision to use a relatively simple CNN architecture in our experiments was made to demonstrate how readily CNNs could be applied to our task, but it was also a result of being confined to the VRAM capacity of the available hardware; introducing additional layers, connections, or training parameters would potentially increase the hardware requirements.
            In the near term, performance gains could be achieved through improved software engineering. In the long term, improvements may come in the form of cost reduction and hardware advances, which would certainly make exploring deeper, wider, or more sophisticated 4D CNN architectures, similar to the many well-established options that are available in lower-dimensions~\citep{Sim14,Sze15,He16}, more tractable; the hope would be to determine which architectures cope best with topological applications, such as estimating Betti numbers. 
            The application of TDA to the network architecture itself may offer some insight into this line of enquiry, similarly as demonstrated in~\citep{Car20}.

        \subsection{Future work}\label{subsec:future_work}
            As previously alluded to, future efforts could extend the results that are presented in this work in several immediate directions.
            The downscaling results could be expanded by implementing and then exploring the use of the 4D equivalents of other image downscaling algorithms, such as those that were mentioned in Section~\ref{sec:background}.
            The downscaling approach that we took in this work could also be compared and combined with \textit{multi-view} methods, which employ lower-dimensional representations of data that are produced by gathering lower-dimensional images from different angles (perspectives), or by projecting 4D data onto a lower-dimensional space; this could potentially offer some speed or memory advantage. This would contrast \textit{volumetric} approaches that process 4D data using 4D operations, such as those applied through a 4D CNN (see Section I.B of~\citet{Cao20} for a discussion about multi-view and volumetric approaches).
            In the 3D setting, \citet{Su15} demonstrate how CAD and voxel data may be projected onto two dimensions by using an `outside' perspective to capture several images of the data from different angles in a similar way to taking X-ray scans of a subject along three orthogonal axes.
            Alternatively, \citet{Shi15} project 3D samples outwards from their centre onto an annulus that wraps around the object.
            Similar ideas have been employed by~\citet{Qi16} and~\citet{Kan18}.
            Nevertheless, \citet{Wan19} suggest that standard volumetric approaches may be more capable of gathering information when compared to multi-view models, and argue that this is because multi-view strategies often fail to encode information from different views. It may, however, be easier to implement larger models in lower dimensions or exploit pre-trained models by fine-tuning~\citep{Rus15, Bro16}, which could potentially be useful when handling large 4D data.

            More generally, several other matters also require deeper investigation in order to fully assess the capabilities of CNNs in estimating topology. 
            For one, the task of modelling a real-world dataset for the purpose of training a CNN would need to be addressed, particularly when it would not be possible to analyse and label real-world data or collect enough data for the purpose of training a CNN. In this case, a CNN would be prepared with a synthetic dataset, which has been designed to exhibit similar features or a comparable distribution to the real-world data, before applying it to analyse real-world data.
            Secondly, generating this data would need to be carefully controlled, possibly via some form of topology-preserving algorithm, as this would allow for the data labels that are produced on-the-fly during the sample generation step (Section~\ref{subsec:data_labelling}) to be carried over without alteration. Solutions will likely require a theoretical and practical collaboration between algebraic topology and computer science. Such an interdisciplinary effort would facilitate the extension of the approach to a broader class of 3-manifolds~\citep{AcetoEtAl2024, Martelli2023, Pur20, Thurston1997, GilmerEtAl1983}, support the characterisation of manifolds in image-type data, and inform how these structures can be accurately represented in synthetic datasets. 
            For example, the random deformation of the canonical embeddings that are used in our experiments could be accomplished by developing a 4D version of the repulsive tangent-point energy algorithm that was proposed by~\citet{Yu21A, Yu21B} (although, we concede that this method itself would be computationally expensive and would not protect against the homological consequences of producing deformed cavities with self-intersections).
            These two matters raise the question of whether generating a diverse training dataset from the outset, that is, a dataset that comprises of more complicated features, such as multiple components, links, and connected sums, could be enough to train a CNN that is capable of general homology estimation, without the need to understand any properties of the real-world dataset under consideration.
            Thirdly, while the CNN architectures that have been considered in this present work demonstrate better performance when a cavity-focused training approach is used, a CNN that sees abstract holes more similarly to persistent homology could possibly be developed. Hyperparameters such as the learning rate, loss function, and number and type of layers could be adjusted, and it would be interesting to compare the problem from the perspective of a prediction problem versus a classification problem.
            Potentially, there may also be some benefit in applying (Bayesian) statistical approaches, such as those considered in zero-shot, one-shot, or $N$-shot learning models~\citep{Fe03,Pal09}, which are capable of generalising knowledge to unfamiliar cases after seeing little, or no, training examples without requiring extensive retraining, or where complete training may not be possible due to dataset limitations, or because real-world data may be infinitely-variable (as may be the case with homeomorphic deformations).
        
    \section{Conclusion}\label{sec:conclusion}
        When samples become large, as is typical for 4D data, the hardware requirements to train CNNs with this data can also grow. Similarly, it can become difficult to meet the computational and memory demands of traditional TDA techniques, such as persistent homology. Alleviating these issues in the context of large point-cloud data is an active area of research~\citep{Cao22, Sol22, Moi18, Cha15b}; 
        The results of our study apply to image-type data and run parallel to this line of research.
        
        The results that are presented in this work demonstrate that downscaling and 4D CNNs work well together in the task of estimating the Betti numbers of manifolds in our 4D simulated image-type data; this is shown under two different training approaches, namely, a cavity-focused approach and a corresponding approach with Betti number targets.
        However, it is conceivable that
        the approach would still have computational constraints when it comes to large real-world samples in 4D. 
        Section~\ref{sec:a_study_using_3d} demonstrates that downscaling can also be applied to real-world data before using them in similar 3D experiments. 

        The artifacts that are introduced into the data through downscaling can be seen as a form of noise that impacts its topology. Our results demonstrate that CNNs possess an additional—and in this case, superior—ability to handle noise beyond that of persistent homology. This is remarkable, given that persistent homology is inherently designed to manage certain levels of noise as they may occur in applications.
        
        Future research could expand on more advanced image-type data generation techniques that would produce more diverse-looking datasets. When coupled with more sophisticated CNN architectures and more powerful hardware, producing even more capable computer vision-based solutions in 4D may be possible.
        Several aspects should make this new line of research and its future development attractive to the graphics community. This includes the outlook to extend graphics and computer vision to 4D, and the possible use of graphics techniques for synthesising training data that may afford 4D real-world applications, for example, in the medical domain.

\begin{acks}
    This research was supported by the Australian Government through the ARC's Discovery Projects funding scheme 
    (DP210103304 `Estimating the Topology of Low-Dimensional Data Using Deep Neural Networks'). 
    The views expressed herein are those of the authors and are not necessarily those of the Australian Government or Australian Research Council. The first author was supported by a PhD scholarship from the University of Newcastle associated with this grant.
    The authors are grateful to Thomas Fiedler for providing a dataset of a CT scan of a manufactured metallic syntactic foam~\citep{Fie20} from which the supplementary 3D data was derived.
\end{acks}

    \bibliographystyle{ACM-Reference-Format}
    \bibliography{ACM_TOG}

    \appendix
    
    \section*{Appendices}
        The following appendices provide further discussion on some of the topics that are introduced in the body of this work. The appendices also provide further context by summarising some material of the workshop paper by~\citet{Han23}.
    
        \section{4D Camera}\label{app:4d_camera}
            
            \begin{figure*}[ht!]
            \begin{minipage}[b]{\textwidth}
                \begin{equation}\label{eqn:convolution}
                   y[c_{out},m,n,o,p] = \sum_{c_{in} \in C_{in}} \sum_{l=-\infty}^{\infty} \sum_{k=-\infty}^{\infty} \sum_{j=-\infty}^{\infty} \sum_{i=-\infty}^{\infty} x[c_{in}, i,j,k,l] \cdot f[c_{out}, c_{in}, m+i,n+j,o+k,p+l]
                \end{equation}
            \end{minipage}
            \end{figure*}
            The neural networks that were used in this work were implemented using custom 4D convolution and pooling layers, which were developed using the PyTorch machine learning framework~\citep{Pas19}. The development and testing of this software is discussed in~\citep{Han23},
            however, the relevant results are summarised here.
    
            \begin{figure*}[ht!]
                \centering
                \definecolor{qqqqff}{rgb}{0.3333333333333333,0.3333333333333333,0.3333333333333333}
                \begin{tikzpicture}[line cap=round,line join=round,>=triangle 45,x=.7cm,y=.7cm]
                	\clip(-6.,-4.5) rectangle (16.,5.5);
                	\fill[line width=.5pt,color=qqqqff,fill=blue,fill opacity=0.47999998927116394] (2.4,1.2) -- (2.,1.) -- (1.,1.) -- (1.,2.) -- (1.4,2.2) -- (2.4,2.2) -- cycle;
                	\fill[line width=.5pt,color=qqqqff,fill=blue,fill opacity=0.47999998927116394] (2.2,2.4) -- (2.2,2.6) -- (2.6,2.8) -- (3.6,2.8) -- (3.6,1.8) -- (3.2,1.6) -- (2.8,1.6) -- (2.8,2.4) -- cycle;
                	\fill[line width=.5pt,dash pattern=on 3pt off 3pt,color=qqqqff,fill=blue,fill opacity=0.1899999976158142] (-4.,-3.) -- (-4.,-2.) -- (-3.6,-1.8) -- (-2.6,-1.8) -- (-2.6,-2.8) -- (-3.,-3.) -- cycle;
                	\fill[line width=.5pt,dash pattern=on 3pt off 3pt,color=qqqqff,fill=blue,fill opacity=0.1899999976158142] (-2.8,-1.4) -- (-2.4,-1.2) -- (-1.4,-1.2) -- (-1.4,-2.2) -- (-1.8,-2.4) -- (-2.2,-2.4) -- (-2.2,-1.6) -- (-2.8,-1.6) -- cycle;
                	\fill[line width=.5pt,dash pattern=on 3pt off 3pt,color=qqqqff,fill=blue,fill opacity=0.1899999976158142] (1.,-3.) -- (1.,-2.) -- (1.4,-1.8) -- (2.4,-1.8) -- (2.4,-2.8) -- (2.,-3.) -- cycle;
                	\fill[line width=.5pt,dash pattern=on 3pt off 3pt,color=qqqqff,fill=blue,fill opacity=0.1899999976158142] (2.2,-1.4) -- (2.6,-1.2) -- (3.6,-1.2) -- (3.6,-2.2) -- (3.2,-2.4) -- (2.8,-2.4) -- (2.8,-1.6) -- (2.2,-1.6) -- cycle;
                	\fill[line width=.5pt,color=qqqqff,fill=orange,fill opacity=0.47999998927116394] (9.1,1.6) -- (9.1,2.1012039945813106) -- (9.299168557005423,2.2003886099783525) -- (9.79980569501089,2.200388609978351) -- (9.799870567450139,1.6994551579091874) -- (9.601436464216809,1.6) -- cycle;
                	\fill[line width=.5pt,dash pattern=on 3pt off 3pt,color=qqqqff,fill=orange,fill opacity=0.1899999976158142] (9.1,-1.9) -- (9.3,-1.8) -- (9.8,-1.8) -- (9.8,-2.3) -- (9.6,-2.4) -- (9.1,-2.4) -- cycle;
                	\fill[line width=.5pt,color=qqqqff,fill=blue,fill opacity=0.47999998927116394] (-4.,2.) -- (-3.6,2.2) -- (-2.6,2.2) -- (-2.6,1.2) -- (-3.,1.) -- (-4.,1.) -- cycle;
                	\fill[line width=.5pt,color=qqqqff,fill=blue,fill opacity=0.47999998927116394] (-2.8,2.4) -- (-2.8,2.6) -- (-2.4,2.8) -- (-1.4,2.8) -- (-1.4,1.8) -- (-1.8,1.6) -- (-2.2,1.6) -- (-2.2,2.4) -- cycle;
                	\draw [line width=.5pt] (0.,0.)-- (2.,0.);
                	\draw [line width=.5pt] (2.,0.)-- (2.,2.);
                	\draw [line width=.5pt] (2.,2.)-- (0.,2.);
                	\draw [line width=.5pt] (0.,2.)-- (0.,0.);
                	\draw [line width=.5pt] (0.,1.5)-- (2.,1.5);
                	\draw [line width=.5pt] (0.,1.)-- (2.,1.);
                	\draw [line width=.5pt] (0.,0.5)-- (2.,0.5);
                	\draw [line width=.5pt] (0.5,2.)-- (0.5,0.);
                	\draw [line width=.5pt] (1.,2.)-- (1.,0.);
                	\draw [line width=.5pt] (1.5,2.)-- (1.5,0.);
                	\draw [line width=.5pt] (0.,2.)-- (0.8,2.4);
                	\draw [line width=.5pt] (2.,2.)-- (2.8,2.4);
                	\draw [line width=.5pt] (0.8,2.4)-- (2.8,2.4);
                	\draw [line width=.5pt] (0.5,2.)-- (1.3,2.4);
                	\draw [line width=.5pt] (1.,2.)-- (1.8,2.4);
                	\draw [line width=.5pt] (1.5,2.)-- (2.3,2.4);
                	\draw [line width=.5pt] (0.2,2.1)-- (2.2,2.1);
                	\draw [line width=.5pt] (0.4,2.2)-- (2.4,2.2);
                	\draw [line width=.5pt] (0.6,2.3)-- (2.6,2.3);
                	\draw [line width=.5pt] (2.,0.)-- (2.8,0.4);
                	\draw [line width=.5pt] (2.8,0.4)-- (2.8,2.4);
                	\draw [line width=.5pt] (2.6,2.3)-- (2.6,0.3);
                	\draw [line width=.5pt] (2.4,2.2)-- (2.4,0.2);
                	\draw [line width=.5pt] (2.2,2.1)-- (2.2,0.1);
                	\draw [line width=.5pt] (2.,1.5)-- (2.8,1.9);
                	\draw [line width=.5pt] (2.,1.)-- (2.8,1.4);
                	\draw [line width=.5pt] (2.,0.5)-- (2.8,0.9);
                	\draw [line width=.5pt] (1.2,2.6)-- (1.2,2.4);
                	\draw [line width=.5pt] (1.2,2.6)-- (2.,3.);
                	\draw [line width=.5pt] (2.,3.)-- (4.,3.);
                	\draw [line width=.5pt] (4.,3.)-- (4.,1.);
                	\draw [line width=.5pt] (1.2,2.6)-- (3.2,2.6);
                	\draw [line width=.5pt] (3.2,2.6)-- (4.,3.);
                	\draw [line width=.5pt] (3.2,2.6)-- (3.2,0.6);
                	\draw [line width=.5pt] (3.2,0.6)-- (4.,1.);
                	\draw [line width=.5pt] (3.2,0.6)-- (2.8,0.6);
                	\draw [line width=.5pt] (5.2,3.6)-- (5.2,1.6);
                	\draw [line width=.5pt] (4.4,3.2)-- (4.4,1.2);
                	\draw [line width=.5pt] (4.4,1.2)-- (4.,1.2);
                	\draw [line width=.5pt] (4.4,1.2)-- (5.2,1.6);
                	\draw [line width=.5pt] (4.4,3.2)-- (5.2,3.6);
                	\draw [line width=.5pt] (5.6,3.8)-- (5.6,1.8);
                	\draw [line width=.5pt] (5.6,1.8)-- (5.2,1.8);
                	\draw [line width=.5pt] (1.6,2.8)-- (3.6,2.8);
                	\draw [line width=.5pt] (1.7,2.6)-- (1.7,2.4);
                	\draw [line width=.5pt] (2.2,2.6)-- (2.2,2.4);
                	\draw [line width=.5pt] (2.7,2.6)-- (2.7,2.4);
                	\draw [line width=.5pt] (1.7,2.6)-- (2.5,3.);
                	\draw [line width=.5pt] (2.2,2.6)-- (3.,3.);
                	\draw [line width=.5pt] (2.7,2.6)-- (3.5,3.);
                	\draw [line width=.5pt] (3.2,1.1)-- (2.8,1.1);
                	\draw [line width=.5pt] (3.2,1.6)-- (2.8,1.6);
                	\draw [line width=.5pt] (3.2,2.1)-- (2.8,2.1);
                	\draw [line width=.5pt] (3.2,2.1)-- (4.,2.5);
                	\draw [line width=.5pt] (3.2,1.6)-- (4.,2.);
                	\draw [line width=.5pt] (3.2,1.1)-- (4.,1.5);
                	\draw [line width=.5pt] (3.4,2.7)-- (3.4,0.7);
                	\draw [line width=.5pt] (3.6,2.8)-- (3.6,0.8);
                	\draw [line width=.5pt] (3.8,2.9)-- (3.8,0.9);
                	\draw [line width=.5pt] (4.4,1.7)-- (4.,1.7);
                	\draw [line width=.5pt] (4.4,1.7)-- (5.2,2.1);
                	\draw [line width=.5pt] (4.,2.2)-- (4.4,2.2);
                	\draw [line width=.5pt] (4.4,2.2)-- (5.2,2.6);
                	\draw [line width=.5pt] (4.,2.7)-- (4.4,2.7);
                	\draw [line width=.5pt] (4.4,2.7)-- (5.2,3.1);
                	\draw [line width=.5pt] (3.9,3.2)-- (3.9,3.);
                	\draw [line width=.5pt] (3.4,3.2)-- (3.4,3.);
                	\draw [line width=.5pt] (2.9,3.2)-- (2.9,3.);
                	\draw [line width=.5pt] (3.4,3.2)-- (4.2,3.6);
                	\draw [line width=.5pt] (3.9,3.2)-- (4.7,3.6);
                	\draw [line width=.5pt] (2.6,3.3)-- (4.6,3.3);
                	\draw [line width=.5pt] (5.2,2.3)-- (5.6,2.3);
                	\draw [line width=.5pt] (5.6,2.3)-- (6.4,2.7);
                	\draw [line width=.5pt] (5.2,2.8)-- (5.6,2.8);
                	\draw [line width=.5pt] (5.6,2.8)-- (6.4,3.2);
                	\draw [line width=.5pt] (5.2,3.3)-- (5.6,3.3);
                	\draw [line width=.5pt] (5.6,3.3)-- (6.4,3.7);
                	\draw [line width=.5pt] (5.1,3.8)-- (5.1,3.6);
                	\draw [line width=.5pt] (5.1,3.8)-- (5.9,4.2);
                	\draw [line width=.5pt] (4.6,3.8)-- (4.6,3.6);
                	\draw [line width=.5pt] (4.6,3.8)-- (5.4,4.2);
                	\draw [line width=.5pt] (4.1,3.6)-- (4.1,3.8);
                	\draw [line width=.5pt] (4.1,3.8)-- (4.9,4.2);
                	\draw [line width=.5pt] (6.2,2.1)-- (6.2,4.1);
                	\draw [line width=.5pt] (5.8,3.9)-- (3.8,3.9);
                	\draw [line width=.5pt] (6.,4.)-- (4.,4.);
                	\draw [line width=.5pt] (6.2,4.1)-- (4.2,4.1);
                	\draw [line width=.5pt] (-3.,0.)-- (-3.,2.);
                	\draw [line width=.5pt] (-3.,2.)-- (-5.,2.);
                	\draw [line width=.5pt] (-5.,2.)-- (-5.,0.);
                	\draw [line width=.5pt] (-5.,0.)-- (-3.,0.);
                	\draw [line width=.5pt] (-3.,0.)-- (-2.2,0.4);
                	\draw [line width=.5pt] (-3.,2.)-- (-2.2,2.4);
                	\draw [line width=.5pt] (-2.2,2.4)-- (-2.2,0.4);
                	\draw [line width=.5pt] (-5.,2.)-- (-4.2,2.4);
                	\draw [line width=.5pt] (-4.2,2.4)-- (-2.2,2.4);
                	\draw [line width=.5pt] (0.6,3.8)-- (1.4,4.2);
                	\draw [line width=.5pt] (0.6,3.8)-- (0.6,2.3);
                	\draw [line width=.5pt] (0.6,3.8)-- (-1.4,3.8);
                	\draw [line width=.5pt] (-1.4,3.8)-- (-0.6,4.2);
                	\draw [line width=.5pt] (-0.6,4.2)-- (1.4,4.2);
                	\draw [line width=.5pt] (2.4,3.2)-- (3.2,3.6);
                	\draw [line width=.5pt] (2.4,3.2)-- (2.4,3.);
                	\draw [line width=.5pt] (2.4,3.2)-- (4.4,3.2);
                	\draw [line width=.5pt] (2.9,3.2)-- (3.7,3.6);
                	\draw [line width=.5pt] (1.8,2.9)-- (3.8,2.9);
                	\draw [line width=.5pt] (3.4,2.7)-- (1.4,2.7);
                	\draw [line width=.5pt] (4.6,1.3)-- (4.6,3.3);
                	\draw [line width=.5pt] (4.8,1.4)-- (4.8,3.4);
                	\draw [line width=.5pt] (5.,1.5)-- (5.,3.5);
                	\draw [line width=.5pt] (3.2,3.6)-- (5.2,3.6);
                	\draw [line width=.5pt] (2.8,3.4)-- (4.8,3.4);
                	\draw [line width=.5pt] (3.,3.5)-- (5.,3.5);
                	\draw [line width=.5pt] (3.6,3.8)-- (4.4,4.2);
                	\draw [line width=.5pt] (3.6,3.8)-- (5.6,3.8);
                	\draw [line width=.5pt] (3.6,3.8)-- (3.6,3.6);
                	\draw [line width=.5pt] (5.6,3.8)-- (6.4,4.2);
                	\draw [line width=.5pt] (6.4,4.2)-- (4.4,4.2);
                	\draw [line width=.5pt] (6.,4.)-- (6.,2.);
                	\draw [line width=.5pt] (5.8,3.9)-- (5.8,1.9);
                	\draw [line width=.5pt] (5.6,1.8)-- (6.4,2.2);
                	\draw [line width=.5pt] (6.4,4.2)-- (6.4,2.2);
                	\draw [line width=.5pt] (1.4,4.2)-- (1.4,2.7);
                	\draw [line width=.5pt] (0.,-2.)-- (2.,-2.);
                	\draw [line width=.5pt] (0.,-2.)-- (0.,-4.);
                	\draw [line width=.5pt] (0.,-4.)-- (2.,-4.);
                	\draw [line width=.5pt] (2.,-4.)-- (2.,-2.);
                	\draw [line width=.5pt] (2.,-2.)-- (2.8,-1.6);
                	\draw [line width=.5pt] (0.,-2.)-- (0.8,-1.6);
                	\draw [line width=.5pt] (0.8,-1.6)-- (2.8,-1.6);
                	\draw [line width=.5pt] (2.,-4.)-- (2.8,-3.6);
                	\draw [line width=.5pt] (2.8,-3.6)-- (2.8,-1.6);
                	\draw [line width=.5pt] (-3.,-2.)-- (-5.,-2.);
                	\draw [line width=.5pt] (-5.,-2.)-- (-5.,-4.);
                	\draw [line width=.5pt] (-5.,-4.)-- (-3.,-4.);
                	\draw [line width=.5pt] (-3.,-4.)-- (-3.,-2.);
                	\draw [line width=.5pt] (-3.,-2.)-- (-2.2,-1.6);
                	\draw [line width=.5pt] (-5.,-2.)-- (-4.2,-1.6);
                	\draw [line width=.5pt] (-4.2,-1.6)-- (-2.2,-1.6);
                	\draw [line width=.5pt] (-3.,-4.)-- (-2.2,-3.6);
                	\draw [line width=.5pt] (-2.2,-3.6)-- (-2.2,-1.6);
                	\draw [line width=.5pt] (5.6,-0.2)-- (6.4,0.2);
                	\draw [line width=.5pt] (5.6,-0.2)-- (3.6,-0.2);
                	\draw [line width=.5pt] (3.6,-0.2)-- (4.4,0.2);
                	\draw [line width=.5pt] (4.4,0.2)-- (6.4,0.2);
                	\draw [line width=.5pt] (5.6,-0.2)-- (5.6,-2.2);
                	\draw [line width=.5pt] (5.6,-2.2)-- (6.4,-1.8);
                	\draw [line width=.5pt] (6.4,-1.8)-- (6.4,0.2);
                	\draw [line width=.5pt] (-1.4,-0.2)-- (0.6,-0.2);
                	\draw [line width=.5pt] (0.6,-0.2)-- (0.6,-1.7);
                	\draw [line width=.5pt] (-1.4,-0.2)-- (-0.6,0.2);
                	\draw [line width=.5pt] (-0.6,0.2)-- (0.,0.2);
                	\draw [line width=.5pt] (0.6,-0.2)-- (1.,0.);
                	
                	\draw [line width=.5pt] (-5.6,2.2)-- (-2.,4.);
                	\draw [line width=.5pt] (0.6,4.6)-- (5.2,4.6);
                	\draw [line width=.5pt] (13.,3.)-- (13.,-1.5);
                	\draw [line width=.5pt] (-1.8,2.6)-- (-1.,3.);
                	\draw [line width=.5pt] (-1.,3.)-- (-3.,3.);
                	\draw [line width=.5pt] (-1.8,2.6)-- (-1.8,0.6);
                	\draw [line width=.5pt] (-1.8,0.6)-- (-2.2,0.6);
                	\draw [line width=.5pt] (-1.8,0.6)-- (-1.,1.);
                	\draw [line width=.5pt] (-1.,1.)-- (-1.,3.);
                	\draw [line width=.5pt] (-2.6,3.2)-- (-1.8,3.6);
                	\draw [line width=.5pt] (-0.6,3.2)-- (0.2,3.6);
                	\draw [line width=.5pt] (0.2,3.6)-- (-1.8,3.6);
                	\draw [line width=.5pt] (-2.6,3.2)-- (-0.6,3.2);
                	\draw [line width=.5pt] (-0.6,1.2)-- (-0.6,3.2);
                	\draw [line width=.5pt] (-0.6,1.2)-- (-1.,1.2);
                	\draw [line width=.5pt] (-0.6,1.2)-- (0.,1.5);
                	\draw [line width=.5pt] (0.2,2.1)-- (0.2,3.6);
                	\draw [line width=.5pt] (-2.6,3.2)-- (-2.6,3.);
                	\draw [line width=.5pt] (-1.4,3.8)-- (-1.4,3.6);
                	\draw [line width=.5pt] (-3.,3.)-- (-3.8,2.6);
                	\draw [line width=.5pt] (-3.8,2.6)-- (-3.8,2.4);
                	\draw [line width=.5pt] (-3.8,2.6)-- (-1.8,2.6);
                	\draw [line width=.5pt] (-3.8,-1.4)-- (-3.,-1.);
                	\draw [line width=.5pt] (-1.8,-1.4)-- (-1.,-1.);
                	\draw [line width=.5pt] (-1.,-1.)-- (-3.,-1.);
                	\draw [line width=.5pt] (-3.8,-1.4)-- (-1.8,-1.4);
                	\draw [line width=.5pt] (-3.8,-1.4)-- (-3.8,-1.6);
                	\draw [line width=.5pt] (-1.8,-3.4)-- (-1.,-3.);
                	\draw [line width=.5pt] (-1.,-3.)-- (-1.,-1.);
                	\draw [line width=.5pt] (-1.8,-1.4)-- (-1.8,-3.4);
                	\draw [line width=.5pt] (-1.8,-3.4)-- (-2.2,-3.4);
                	\draw [line width=.5pt] (-2.6,-0.8)-- (-1.8,-0.4);
                	\draw [line width=.5pt] (-0.6,-0.8)-- (0.2,-0.4);
                	\draw [line width=.5pt] (0.2,-0.4)-- (-1.8,-0.4);
                	\draw [line width=.5pt] (-0.6,-0.8)-- (-2.6,-0.8);
                	\draw [line width=.5pt] (-0.6,-2.8)-- (0.,-2.5);
                	\draw [line width=.5pt] (0.2,-0.4)-- (0.2,-1.9);
                	\draw [line width=.5pt] (-0.6,-2.8)-- (-0.6,-0.8);
                	\draw [line width=.5pt] (-0.6,-2.8)-- (-1.,-2.8);
                	\draw [line width=.5pt] (-2.6,-0.8)-- (-2.6,-1.);
                	\draw [line width=.5pt] (-1.4,-0.2)-- (-1.4,-0.4);
                	\draw [line width=.5pt,dash pattern=on 3pt off 3pt,color=qqqqff] (-4.,-3.)-- (-4.,-2.);
                	\draw [line width=.5pt,dash pattern=on 3pt off 3pt,color=qqqqff] (-4.,-2.)-- (-3.6,-1.8);
                	\draw [line width=.5pt,dash pattern=on 3pt off 3pt,color=qqqqff] (-3.6,-1.8)-- (-2.6,-1.8);
                	\draw [line width=.5pt,dash pattern=on 3pt off 3pt,color=qqqqff] (-2.6,-1.8)-- (-2.6,-2.8);
                	\draw [line width=.5pt,dash pattern=on 3pt off 3pt,color=qqqqff] (-2.6,-2.8)-- (-3.,-3.);
                	\draw [line width=.5pt,dash pattern=on 3pt off 3pt,color=qqqqff] (-3.,-3.)-- (-4.,-3.);
                	\draw [line width=.5pt,dash pattern=on 3pt off 3pt,color=qqqqff] (-2.8,-1.4)-- (-2.4,-1.2);
                	\draw [line width=.5pt,dash pattern=on 3pt off 3pt,color=qqqqff] (-2.4,-1.2)-- (-1.4,-1.2);
                	\draw [line width=.5pt,dash pattern=on 3pt off 3pt,color=qqqqff] (-1.4,-1.2)-- (-1.4,-2.2);
                	\draw [line width=.5pt,dash pattern=on 3pt off 3pt,color=qqqqff] (-1.4,-2.2)-- (-1.8,-2.4);
                	\draw [line width=.5pt,dash pattern=on 3pt off 3pt,color=qqqqff] (-1.8,-2.4)-- (-2.2,-2.4);
                	\draw [line width=.5pt,dash pattern=on 3pt off 3pt,color=qqqqff] (-2.2,-2.4)-- (-2.2,-1.6);
                	\draw [line width=.5pt,dash pattern=on 3pt off 3pt,color=qqqqff] (-2.2,-1.6)-- (-2.8,-1.6);
                	\draw [line width=.5pt,dash pattern=on 3pt off 3pt,color=qqqqff] (-2.8,-1.6)-- (-2.8,-1.4);
                	\draw [line width=.5pt] (1.2,-1.4)-- (2.,-1.);
                	\draw [line width=.5pt] (3.2,-1.4)-- (4.,-1.);
                	\draw [line width=.5pt] (4.,-1.)-- (2.,-1.);
                	\draw [line width=.5pt] (1.2,-1.4)-- (3.2,-1.4);
                	\draw [line width=.5pt] (1.2,-1.4)-- (1.2,-1.6);
                	\draw [line width=.5pt] (4.,-1.)-- (4.,-3.);
                	\draw [line width=.5pt] (3.2,-1.4)-- (3.2,-3.4);
                	\draw [line width=.5pt] (3.2,-3.4)-- (2.8,-3.4);
                	\draw [line width=.5pt] (3.2,-3.4)-- (4.,-3.);
                	\draw [line width=.5pt] (2.4,-0.8)-- (3.2,-0.4);
                	\draw [line width=.5pt] (4.4,-0.8)-- (5.2,-0.4);
                	\draw [line width=.5pt] (5.2,-0.4)-- (3.2,-0.4);
                	\draw [line width=.5pt] (2.4,-0.8)-- (4.4,-0.8);
                	\draw [line width=.5pt] (2.4,-0.8)-- (2.4,-1.);
                	\draw [line width=.5pt] (3.6,-0.2)-- (3.6,-0.4);
                	\draw [line width=.5pt] (4.4,-2.8)-- (5.2,-2.4);
                	\draw [line width=.5pt] (4.4,-0.8)-- (4.4,-2.8);
                	\draw [line width=.5pt] (5.2,-2.4)-- (5.2,-0.4);
                	\draw [line width=.5pt] (4.4,-2.8)-- (4.,-2.8);
                	\draw [line width=.5pt] (5.6,-2.2)-- (5.2,-2.2);
                	\draw [line width=.5pt,dash pattern=on 3pt off 3pt,color=qqqqff] (1.,-3.)-- (1.,-2.);
                	\draw [line width=.5pt,dash pattern=on 3pt off 3pt,color=qqqqff] (1.,-2.)-- (1.4,-1.8);
                	\draw [line width=.5pt,dash pattern=on 3pt off 3pt,color=qqqqff] (1.4,-1.8)-- (2.4,-1.8);
                	\draw [line width=.5pt,dash pattern=on 3pt off 3pt,color=qqqqff] (2.4,-1.8)-- (2.4,-2.8);
                	\draw [line width=.5pt,dash pattern=on 3pt off 3pt,color=qqqqff] (2.4,-2.8)-- (2.,-3.);
                	\draw [line width=.5pt,dash pattern=on 3pt off 3pt,color=qqqqff] (2.,-3.)-- (1.,-3.);
                	\draw [line width=.5pt,dash pattern=on 3pt off 3pt,color=qqqqff] (2.2,-1.4)-- (2.6,-1.2);
                	\draw [line width=.5pt,dash pattern=on 3pt off 3pt,color=qqqqff] (2.6,-1.2)-- (3.6,-1.2);
                	\draw [line width=.5pt,dash pattern=on 3pt off 3pt,color=qqqqff] (3.6,-1.2)-- (3.6,-2.2);
                	\draw [line width=.5pt,dash pattern=on 3pt off 3pt,color=qqqqff] (3.6,-2.2)-- (3.2,-2.4);
                	\draw [line width=.5pt,dash pattern=on 3pt off 3pt,color=qqqqff] (3.2,-2.4)-- (2.8,-2.4);
                	\draw [line width=.5pt,dash pattern=on 3pt off 3pt,color=qqqqff] (2.8,-2.4)-- (2.8,-1.6);
                	\draw [line width=.5pt,dash pattern=on 3pt off 3pt,color=qqqqff] (2.8,-1.6)-- (2.2,-1.6);
                	\draw [line width=.5pt,dash pattern=on 3pt off 3pt,color=qqqqff] (2.2,-1.6)-- (2.2,-1.4);
                	\draw [line width=.5pt] (1.4,0.)-- (1.4,-1.3);
                	\draw [line width=.5pt] (8.10143646421681,2.1012039945813106)-- (8.70143646421681,2.4012039945813104);
                	\draw [line width=.5pt] (8.10143646421681,0.6012039945813099)-- (8.10143646421681,2.1012039945813106);
                	\draw [line width=.5pt] (8.70143646421681,2.4012039945813104)-- (10.20143646421681,2.4012039945813104);
                	\draw [line width=.5pt] (10.20143646421681,2.4012039945813104)-- (9.601436464216809,2.1012039945813106);
                	\draw [line width=.5pt] (9.601436464216809,2.1012039945813106)-- (8.10143646421681,2.1012039945813106);
                	\draw [line width=.5pt] (9.601436464216809,0.6012039945813099)-- (10.20143646421681,0.90120399458131);
                	\draw [line width=.5pt] (9.601436464216809,0.6012039945813099)-- (9.601436464216809,2.1012039945813106);
                	\draw [line width=.5pt] (10.20143646421681,0.90120399458131)-- (10.20143646421681,2.4012039945813104);
                	\draw [line width=.5pt] (9.601436464216809,0.6012039945813099)-- (8.10143646421681,0.6012039945813099);
                	\draw [line width=.5pt] (9.30143646421681,2.701203994581311)-- (9.90143646421681,3.001203994581311);
                	\draw [line width=.5pt] (9.30143646421681,2.701203994581311)-- (9.30143646421681,2.4012039945813104);
                	\draw [line width=.5pt] (9.30143646421681,2.701203994581311)-- (10.80143646421681,2.701203994581311);
                	\draw [line width=.5pt] (10.80143646421681,2.701203994581311)-- (11.40143646421681,3.001203994581311);
                	\draw [line width=.5pt] (11.40143646421681,3.001203994581311)-- (9.90143646421681,3.001203994581311);
                	\draw [line width=.5pt] (10.80143646421681,2.701203994581311)-- (10.80143646421681,1.2012039945813102);
                	\draw [line width=.5pt] (10.80143646421681,1.2012039945813102)-- (10.20143646421681,1.2012039945813102);
                	\draw [line width=.5pt] (10.80143646421681,1.2012039945813102)-- (11.4,1.5);
                	\draw [line width=.5pt] (11.4,1.5)-- (11.40143646421681,3.001203994581311);
                	\draw [line width=.5pt] (10.501436464216809,3.3012039945813108)-- (11.101436464216809,3.6012039945813106);
                	\draw [line width=.5pt] (10.501436464216809,3.3012039945813108)-- (10.501436464216809,3.001203994581311);
                	\draw [line width=.5pt] (11.4,1.8)-- (12.,1.8);
                	\draw [line width=.5pt] (12.,1.8)-- (12.6,2.1);
                	\draw [line width=.5pt] (12.001436464216809,3.3012039945813108)-- (12.601436464216809,3.6012039945813106);
                	\draw [line width=.5pt] (12.601436464216809,3.6012039945813106)-- (12.6,2.1);
                	\draw [line width=.5pt] (12.601436464216809,3.6012039945813106)-- (11.101436464216809,3.6012039945813106);
                	\draw [line width=.5pt] (12.001436464216809,3.3012039945813108)-- (10.501436464216809,3.3012039945813108);
                	\draw [line width=.5pt] (12.001436464216809,3.3012039945813108)-- (12.,1.8);
                	\draw [line width=.5pt] (8.6,2.1)-- (9.2,2.4012039945813104);
                	\draw [line width=.5pt] (9.1,2.1012039945813106)-- (9.7,2.4);
                	\draw [line width=.5pt] (9.601436464216809,1.6)-- (10.2,1.9);
                	\draw [line width=.5pt] (9.601436464216809,1.1)-- (10.2,1.4);
                	\draw [line width=.5pt] (8.29980569501089,2.2003886099783507)-- (9.79980569501089,2.200388609978351);
                	\draw [line width=.5pt] (8.49980569501089,2.3003886099783504)-- (10.,2.3);
                	\draw [line width=.5pt] (10.,2.3)-- (10.,0.8);
                	\draw [line width=.5pt] (9.79980569501089,2.200388609978351)-- (9.8,0.7);
                	\draw [line width=.5pt] (8.6,2.1)-- (8.6,0.6012039945813099);
                	\draw [line width=.5pt] (9.1,2.1012039945813106)-- (9.1,0.6012039945813099);
                	\draw [line width=.5pt] (9.601436464216809,1.6)-- (8.1,1.6);
                	\draw [line width=.5pt] (9.601436464216809,1.1)-- (8.1,1.1);
                	
                	\draw [line width=.5pt] (9.8,2.701203994581311)-- (9.8,2.4);
                	\draw [line width=.5pt] (9.8,2.701203994581311)-- (10.4,3.);
                	\draw [line width=.5pt] (10.3,2.701203994581311)-- (10.9,3.);
                	\draw [line width=.5pt] (9.499805695010888,2.8003886099783504)-- (10.999805695010888,2.8003886099783504);
                	\draw [line width=.5pt] (11.199805695010888,2.90038860997835)-- (9.7,2.9);
                	\draw [line width=.5pt] (10.3,2.701203994581311)-- (10.3,1.2);
                	\draw [line width=.5pt] (10.80143646421681,2.2)-- (10.2,2.2);
                	\draw [line width=.5pt] (10.2,1.7)-- (10.8,1.7);
                	\draw [line width=.5pt] (10.8,1.7)-- (11.4,2.);
                	\draw [line width=.5pt] (10.80143646421681,2.2)-- (11.4,2.5);
                	\draw [line width=.5pt] (10.999805695010888,2.8003886099783504)-- (10.99987027907508,1.3002598636329785);
                	\draw [line width=.5pt] (11.199935139537542,1.4001299318164901)-- (11.199805695010888,2.90038860997835);
                	\draw [line width=.5pt] (11.,3.3012039945813108)-- (11.6,3.6);
                	\draw [line width=.5pt] (11.,3.3012039945813108)-- (11.,3.001203994581311);
                	\draw [line width=.5pt] (11.5,3.3)-- (11.5,1.8);
                	\draw [line width=.5pt] (12.000478436943531,2.2999995421957626)-- (11.4,2.3);
                	\draw [line width=.5pt] (12.,2.8)-- (11.4,2.8);
                	\draw [line width=.5pt] (10.699805695010788,3.4003886099783003)-- (12.199805695010737,3.4003886099782745);
                	\draw [line width=.5pt] (11.5,3.3)-- (12.1,3.6);
                	\draw [line width=.5pt] (10.9,3.5)-- (12.4,3.5);
                	\draw [line width=.5pt] (12.,2.8)-- (12.6,3.1);
                	\draw [line width=.5pt] (12.199805695010737,3.4003886099782745)-- (12.2,1.9);
                	\draw [line width=.5pt] (12.000478436943531,2.2999995421957626)-- (12.600478436943531,2.599999542195763);
                	\draw [line width=.5pt] (12.4,3.5)-- (12.4,2.);
                	\draw [line width=.5pt] (8.1,-1.9)-- (8.1,-3.4);
                	\draw [line width=.5pt] (8.1,-1.9)-- (9.6,-1.9);
                	\draw [line width=.5pt] (9.6,-1.9)-- (9.6,-3.4);
                	\draw [line width=.5pt] (8.1,-3.4)-- (9.6,-3.4);
                	\draw [line width=.5pt] (9.6,-3.4)-- (10.2,-3.1);
                	\draw [line width=.5pt] (9.6,-1.9)-- (10.2,-1.6);
                	\draw [line width=.5pt] (10.2,-1.6)-- (10.2,-3.1);
                	\draw [line width=.5pt] (8.1,-1.9)-- (8.7,-1.6);
                	\draw [line width=.5pt] (8.7,-1.6)-- (10.2,-1.6);
                	\draw [line width=.5pt,dash pattern=on 3pt off 3pt,color=qqqqff] (9.1,-1.9)-- (9.3,-1.8);
                	\draw [line width=.5pt,dash pattern=on 3pt off 3pt,color=qqqqff] (9.3,-1.8)-- (9.8,-1.8);
                	\draw [line width=.5pt,dash pattern=on 3pt off 3pt,color=qqqqff] (9.8,-1.8)-- (9.8,-2.3);
                	\draw [line width=.5pt,dash pattern=on 3pt off 3pt,color=qqqqff] (9.8,-2.3)-- (9.6,-2.4);
                	\draw [line width=.5pt,dash pattern=on 3pt off 3pt,color=qqqqff] (9.6,-2.4)-- (9.1,-2.4);
                	\draw [line width=.5pt,dash pattern=on 3pt off 3pt,color=qqqqff] (9.1,-2.4)-- (9.1,-1.9);
                	\draw [line width=.5pt] (9.2,-1.6)-- (9.2,-1.3);
                	\draw [line width=.5pt] (9.2,-1.3)-- (9.8,-1.);
                	\draw [line width=.5pt] (9.8,-1.)-- (11.4,-1.);
                	\draw [line width=.5pt] (11.4,-1.)-- (10.8,-1.3);
                	\draw [line width=.5pt] (10.8,-1.3)-- (9.2,-1.3);
                	\draw [line width=.5pt] (10.8,-1.3)-- (10.8,-2.8);
                	\draw [line width=.5pt] (10.8,-2.8)-- (10.2,-2.8);
                	\draw [line width=.5pt] (10.8,-2.8)-- (11.4,-2.5);
                	\draw [line width=.5pt] (11.4,-2.5)-- (11.4,-1.);
                	\draw [line width=.5pt] (10.4,-1.)-- (10.4,-0.7);
                	\draw [line width=.5pt] (10.4,-0.7)-- (11.,-0.4);
                	\draw [line width=.5pt] (11.,-0.4)-- (12.6,-0.4);
                	\draw [line width=.5pt] (12.6,-0.4)-- (12.,-0.7);
                	\draw [line width=.5pt] (10.4,-0.7)-- (12.,-0.7);
                	\draw [line width=.5pt] (12.,-0.7)-- (12.,-2.2);
                	\draw [line width=.5pt] (12.,-2.2)-- (12.6,-1.9);
                	\draw [line width=.5pt] (12.6,-1.9)-- (12.6,-0.4);
                	\draw [line width=.5pt] (12.,-2.2)-- (11.4,-2.2);
                	\draw [line width=.5pt] (-4.5,0.)-- (-4.5,2.);
                	\draw [line width=.5pt] (-4.,2.)-- (-4.,0.);
                	\draw [line width=.5pt] (-3.5,0.)-- (-3.5,2.);
                	\draw [line width=.5pt] (-5.,1.5)-- (-3.,1.5);
                	\draw [line width=.5pt] (-3.,1.)-- (-5.,1.);
                	\draw [line width=.5pt] (-5.,0.5)-- (-3.,0.5);
                	\draw [line width=.5pt] (-4.5,2.)-- (-3.7,2.4);
                	\draw [line width=.5pt] (-4.8,2.1)-- (-2.8,2.1);
                	\draw [line width=.5pt] (-2.6,2.2)-- (-4.6,2.2);
                	\draw [line width=.5pt] (-4.4,2.3)-- (-2.4,2.3);
                	\draw [line width=.5pt] (-4.,2.)-- (-3.2,2.4);
                	\draw [line width=.5pt] (-3.5,2.)-- (-2.7,2.4);
                	\draw [line width=.5pt] (-2.8,2.1)-- (-2.8,0.1);
                	\draw [line width=.5pt] (-2.6,2.2)-- (-2.6,0.2);
                	\draw [line width=.5pt] (-2.4,2.3)-- (-2.4,0.3);
                	\draw [line width=.5pt] (-3.,0.5)-- (-2.2,0.9);
                	\draw [line width=.5pt] (-3.,1.)-- (-2.2,1.4);
                	\draw [line width=.5pt] (-3.,1.5)-- (-2.2,1.9);
                	
                	\draw [line width=.5pt] (-3.3,2.4)-- (-3.3,2.6);
                	\draw [line width=.5pt] (-2.8,2.4)-- (-2.8,2.6);
                	\draw [line width=.5pt] (-2.3,2.4)-- (-2.3,2.6);
                	\draw [line width=.5pt] (-1.8,2.1)-- (-2.2,2.1);
                	\draw [line width=.5pt] (-2.2,1.6)-- (-1.8,1.6);
                	\draw [line width=.5pt] (-2.2,1.1)-- (-1.8,1.1);
                	\draw [line width=.5pt] (-1.8,1.1)-- (-1.,1.5);
                	\draw [line width=.5pt] (-1.6,0.7)-- (-1.6,2.7);
                	\draw [line width=.5pt] (-1.8,1.6)-- (-1.,2.);
                	\draw [line width=.5pt] (-1.4,0.8)-- (-1.4,2.8);
                	\draw [line width=.5pt] (-1.8,2.1)-- (-1.,2.5);
                	\draw [line width=.5pt] (-1.2,0.9)-- (-1.2,2.9);
                	\draw [line width=.5pt] (-1.2,2.9)-- (-3.2,2.9);
                	\draw [line width=.5pt] (-1.5,3.)-- (-2.3,2.6);
                	\draw [line width=.5pt] (-1.4,2.8)-- (-3.4,2.8);
                	\draw [line width=.5pt] (-2.,3.)-- (-2.8,2.6);
                	\draw [line width=.5pt] (-1.6,2.7)-- (-3.6,2.7);
                	\draw [line width=.5pt] (-2.5,3.)-- (-3.3,2.6);
                	
                	\draw [line width=.5pt] (-2.1,3.)-- (-2.1,3.2);
                	\draw [line width=.5pt] (-1.6,3.)-- (-1.6,3.2);
                	\draw [line width=.5pt] (-1.1,3.)-- (-1.1,3.2);
                	\draw [line width=.5pt] (-0.3,3.6)-- (-1.1,3.2);
                	\draw [line width=.5pt] (-2.1,3.2)-- (-1.3,3.6);
                	\draw [line width=.5pt] (-1.6,3.2)-- (-0.8,3.6);
                	\draw [line width=.5pt] (-2.2,3.4)-- (-0.2,3.4);
                	\draw [line width=.5pt] (-2.,3.5)-- (0.,3.5);
                	\draw [line width=.5pt] (-2.4,3.3)-- (-0.4,3.3);
                	\draw [line width=.5pt] (-0.6,2.7)-- (-1.,2.7);
                	\draw [line width=.5pt] (-0.6,2.7)-- (0.2,3.1);
                	\draw [line width=.5pt] (-1.,2.2)-- (-0.6,2.2);
                	\draw [line width=.5pt] (-1.,1.7)-- (-0.6,1.7);
                	\draw [line width=.5pt] (-0.6,1.7)-- (0.,2.);
                	\draw [line width=.5pt] (-0.4,1.3)-- (-0.4,3.3);
                	\draw [line width=.5pt] (-0.2,3.4)-- (-0.2,1.4);
                	\draw [line width=.5pt] (0.,3.5)-- (0.,2.);
                	\draw [line width=.5pt] (-0.6,2.2)-- (0.2,2.6);
                	\draw [line width=.5pt] (-1.2,3.9)-- (0.8,3.9);
                	\draw [line width=.5pt] (1.,4.)-- (-1.,4.);
                	\draw [line width=.5pt] (-0.8,4.1)-- (1.2,4.1);
                	\draw [line width=.5pt] (0.9,4.2)-- (0.1,3.8);
                	\draw [line width=.5pt] (0.4,4.2)-- (-0.4,3.8);
                	\draw [line width=.5pt] (-0.1,4.2)-- (-0.9,3.8);
                	\draw [line width=.5pt] (-0.9,3.8)-- (-0.9,3.6);
                	\draw [line width=.5pt] (-0.4,3.8)-- (-0.4,3.6);
                	\draw [line width=.5pt] (0.1,3.8)-- (0.1,3.6);
                	\draw [line width=.5pt] (0.2,3.3)-- (0.6,3.3);
                	\draw [line width=.5pt] (0.2,2.8)-- (0.6,2.8);
                	\draw [line width=.5pt] (0.2,2.3)-- (0.6,2.3);
                	\draw [line width=.5pt] (0.6,2.3)-- (1.4,2.7);
                	\draw [line width=.5pt] (0.6,2.8)-- (1.4,3.2);
                	\draw [line width=.5pt] (0.6,3.3)-- (1.4,3.7);
                	\draw [line width=.5pt] (1.2,4.1)-- (1.2,2.6);
                	\draw [line width=.5pt] (1.,4.)-- (1.,2.4);
                	\draw [line width=.5pt] (0.8,3.9)-- (0.8,2.4);
                	\draw [line width=.5pt,dash pattern=on 1pt off 2pt] (-2.8,-2.4)-- (1.,-2.4);
                	\draw [line width=.5pt,dash pattern=on 1pt off 2pt] (-1.6,-1.8)-- (0.4,-1.8);
                	\draw [line width=.5pt,dash pattern=on 1pt off 2pt] (-2.8,1.6)-- (1.,1.6);
                	\draw [line width=.5pt,dash pattern=on 1pt off 2pt] (-1.6,2.2)-- (0.4,2.2);
                	\draw [line width=.5pt,dash pattern=on 1pt off 2pt] (9.1,-2.1)-- (8.,-2.1);
                	\draw [line width=.5pt,dash pattern=on 1pt off 2pt] (3.4,-1.8)-- (8.,-2.1);
                	\draw [line width=.5pt,dash pattern=on 1pt off 2pt] (8.,-2.1)-- (2.2,-2.4);
                	\draw [line width=.5pt,dash pattern=on 1pt off 2pt] (9.1,1.9)-- (8.,1.9);
                	\draw [line width=.5pt,dash pattern=on 1pt off 2pt] (3.4,2.2)-- (8.,1.9);
                	\draw [line width=.5pt,dash pattern=on 1pt off 2pt] (2.218007334750181,1.6020962051777348)-- (8.,1.9);
                	\begin{scriptsize}
                		\draw [fill=black,shift={(-5.6,2.2)}] (0,0) ++(0 pt,2.25pt) -- ++(1.9485571585149868pt,-3.375pt)--++(-3.8971143170299736pt,0 pt) -- ++(1.9485571585149868pt,3.375pt);
                		\draw [fill=black,shift={(-2.,4.)},rotate=180] (0,0) ++(0 pt,2.25pt) -- ++(1.9485571585149868pt,-3.375pt)--++(-3.8971143170299736pt,0 pt) -- ++(1.9485571585149868pt,3.375pt);
                		\draw[color=black] (-3.8227062566470726,3.3744008238572683) node {$w$};
                		\draw[color=black] (0.5,5.) node {$c_{in_{0}}$};
                		\draw [fill=black,shift={(5.2,4.6)},rotate=270] (0,0) ++(0 pt,2.25pt) -- ++(1.9485571585149868pt,-3.375pt)--++(-3.8971143170299736pt,0 pt) -- ++(1.9485571585149868pt,3.375pt);
                		\draw[color=black] (5.566605790686001,5.) node {$c_{in_{N-1}}$};
                		\draw[color=black] (2.9,5.) node {Input channels};
                		\draw[color=black] (13.7,3.) node {$c_{out_{0}}$};
                		\draw [fill=black,shift={(13.,-1.5)},rotate=180] (0,0) ++(0 pt,2.25pt) -- ++(1.9485571585149868pt,-3.375pt)--++(-3.8971143170299736pt,0 pt) -- ++(1.9485571585149868pt,3.375pt);
                		\draw[color=black] (14.,-1.296568788697831) node {$c_{out_{M-1}}$};
                		\draw[color=black] (13.9,.8) node {Output};
                		\draw[color=black] (13.9,.5) node {channels};
                	\end{scriptsize}
                \end{tikzpicture}

                \caption{A 4D camera. This example of a 4D convolution shows a filter, which consists of $2^4$ kernels, taking an $N$-channelled $4^4$ sample as input and producing an $M$-channelled $3^4$ feature map. The input is shown on the left side of the figure and the output is shown on the right. The filter is coloured blue and the result of the convolution at its current location is sent to the orange toxel in the feature map. A variation in colour opacity (from top to bottom) is used to signify that the camera is producing distinct output channels. For example, by fixing $c_{out} = c_{out_{0}}$ in Equation~\ref{eqn:convolution}, we would simply have the top (most opaque) row of the input, filter, and feature map that are depicted in this figure.
                }
                \Description{An illustration of a 4D sample undergoing convolution.}
                \label{fig:4d_camera}
            \end{figure*}
            
            For a sample $x$, filter $f$, and output channel $c_{out} \in C_{out}$, the 4D convolution operation was taken as a sum of the convolutions over each input channel $c_{in} \in C_{in}$, as expressed in Equation~\ref{eqn:convolution}. This equation is analogous to the convolution operation in the 2D and 3D setting, where, for example, a 2D RGB image can be considered as 3-channelled 2D data.
            
            The operation of Equation~\ref{eqn:convolution} is illustrated in Figure~\ref{fig:4d_camera}, which shows a filter that consists of $2^4$ kernels taking an $N$-channelled $4^4$ sample as input and producing an $M$-channelled $3^4$ feature map.

            Three approaches to implementing the 4D convolution were investigated, each with user-definable kernel dimensions, padding, stride length, and GPU acceleration compatibility. An optional bias vector $B$ was included, which contained a bias $b_{out}$ for each output channel. Therefore, each approach computed an operation to the effect of $y[c_{out},m,n,o,p] + b_{out}$.
            
            Following a series of pilot tests, an implementation over PyTorch's native 3D convolution operation was selected because it performed well with respect to speed and potential batch size versus the other options. This approach is essentially a rearrangement of Equation~\ref{eqn:convolution}, which is afforded since the sum is finite in practice. This modification results in a sum of 3D convolutions over the remaining dimension (indexed by $l$). A similar approach was used to implement the 4D pooling layers. This software is included in~\citep{Han23supp}.

        \section{Notes on algebraic topology}\label{app_sec:algebraic_topology}
    
            A general introduction to algebraic topology that can serve as background to our approach can be found in the books of~\citet{Ede10} and~\citet{Hatcher2002}.
            Here, we elaborate on some of the concepts that were referenced in Section~\ref{subsec:data_labelling} of the paper.
            
            \subsection{Reduced homology}\label{app_subsec:reduced_homology}
                In contrast to the higher dimensional Betti numbers, $\beta_0$ is often interpreted as the number of connected components, rather than some number of holes. It would seem more consistent if $\beta_0 = 1$ implied the existence of a `gap' between two components; this is the rationale behind using reduced homology, and its application can simplify some calculations. The modification is achieved by introducing an augmentation map into the chain complex that is used in the derivation of homology theory, and the $n$\textsuperscript{th} \emph{reduced} homology group is often denoted $\tilde{H}_n$. The $p$th reduced Betti number is denoted~$\tilde{\beta}_p$, and is analogously defined as $\mathrm{rank} \tilde{H}_p$. The effect of these changes is that $\tilde{\beta}_p = \beta_p$ for all $p > 0$, and $\tilde{\beta}_0 = \beta_0 -1$, as desired.
                The reader is directed to~\citet[Chapter 2]{Hatcher2002} for more.
             
            \subsection{The Mayer-Vietoris sequence}\label{app_subsec:MV_sequence}
                We state two versions of the Mayer-Vietoris sequence in singular homology. A proof can be found in~\citep[Chapter 4.4]{Ede10}, and more discussion can be found in~\citep[Chapter 2.2]{Hatcher2002}.
                
                Let $X$ be a topological space, and $A$ and $B$ be two subspaces whose interiors cover $X$; the interiors of $A$ and $B$ may intersect. The Mayer-Vietoris sequence is a long exact sequence that relates the singular homology groups (with coefficient group $\mathbb{Z}$) of $X$, $A$, $B$, and $A\cap B$ by
            	\begin{equation}
                	\begin{split}
                	    \cdots 
                		\rightarrow H_{n+1}(X) 
                		\xrightarrow{\partial_{n+1}} H_{n}(A \cap B) 
                		\xrightarrow{(i_n, j_n)} H_{n}(A) \oplus H_{n}(B) \\
                		\xrightarrow{k_n-l_n}  H_{n}(X) 
                		\xrightarrow{\partial_{n}} H_{n-1}(A \cap B)
                		\rightarrow 
                		\cdots  
                		\rightarrow H_{0}(A) \oplus H_{0}(B) \\
                		\xrightarrow{k_0-l_0} H_{0}(X)
                		\rightarrow 0,\label{app_eqn:MV}
                	\end{split}
            	\end{equation} 
                where $i : A \cap B \rightarrow  A$, $j : A\cap B \rightarrow B$, $k : A \rightarrow X$, and $l : B \rightarrow X$ are inclusion maps, $\oplus$ denotes the direct sum, and $\partial_n$ denotes the $n$\textsuperscript{th} boundary homomorphism.
                
            	Assuming that the intersection of $A$ and $B$ is not empty, the Mayer-Vietoris sequence for reduced homology is identical to Equation~\ref{app_eqn:MV} for $n>0$, and ends with
            	\begin{equation}
                	\cdots 
                	\rightarrow \tilde{H}_{0}(A \cap B)
                	\xrightarrow{(i_0,j_0)} \tilde{H}_{0}(A) \oplus \tilde{H}_{0}(B) 
                	\xrightarrow{k_0-l_0} \tilde{H}_{0}(X)
                	\rightarrow 0.
                \end{equation}

            \subsection{The K\"{u}nneth theorem}\label{app_subsec:kunneth_theoreem}
                
                The classical statement of the K\"{u}nneth theorem for principal ideal domains, such as any field $\mathbb{F}$, or as in our case, the ring of integers $\mathbb{Z}$, relates the singular homology of two topological spaces $X$ and $Y$ with their product space $X \times Y$. The reader is directed to~\citet[Chapter 3.B]{Hatcher2002} for a review of several versions of this theorem, and an explanation of the $\mathrm{Tor}$ functor.
                				
            	Given a principal ideal domain $R$, and any topological spaces $X$ and $Y$, the K\"{u}nneth theorem states that there are short exact sequences, such that
            	\begin{equation}
                	\begin{split}
                    	0 \rightarrow 
                    	\bigoplus_{i+j=k} {H_i(X;R) \otimes_R H_j(Y;R)} \rightarrow 
                    	H_k(X \times Y;R) \rightarrow \\
                    	\bigoplus_{i+j=k-1}{\mathrm{Tor}^R_1(H_i(X;R), H_j(Y;R))} \rightarrow
                    	0,
                    \end{split}
                \end{equation}
                where $\otimes_R$ denotes the tensor product.
    
            \subsection{Label derivation}\label{app_subsec:label_derivation}
                We demonstrate an application of these ideas by calculating the Betti numbers of $I^4 - (S^1 \times S^1 \times B^2)$.	
                Since the homology groups of the factors of $S^1 \times S^1 \times B^2$ and its boundary $S^1 \times S^1 \times S^1$ are well-known, it can be shown that the $\mathrm{Tor}$ functor components in the K\"unneth theorem's short exact sequences are trivial, which implies the following isomorphisms.
                \begin{align}
            		H_n(S^1 \times S^1 \times B^2) \cong 
            		\begin{cases}
            		\mathbb{Z} & n = 0,2 \\
            		\mathbb{Z}^2 & n = 1 \\
            		0 & \text{otherwise }\\
            		\end{cases}
            	\end{align}
            	\begin{align}
            		H_n(S^1 \times S^1 \times S^1) \cong 
            		\begin{cases}
            		\mathbb{Z} & n = 0,3 \\
            		\mathbb{Z}^3 & n = 1,2 \\
            		0 & \text{otherwise }\\
            		\end{cases}
            	\end{align}
            	
            	We then define an embedding $\varphi : S^1 \times S^1 \times B^2 \rightarrow I^4$, and let $K = \varphi(S^1 \times S^1 \times B^2)$ and $X = I^4 - K$. We also define $Y = K \cup N(K)$, where $N(K)$ is an open neighbourhood of $K$, so that we have $X \cup Y = I^4$, and the homotopy equivalence $X \cap Y \simeq S^1 \times S^1 \times S^1$. The Mayer-Vietoris Sequence is then applied using $X$ and $Y$ in light of the isomorphism between the homology groups of homotopy equivalent spaces in singular homology~\citep[Chapter 2]{Hatcher2002}.
                For the $n=0$ case, we apply the reduced Mayer-Vietoris sequence, and collectively, these results imply that
                \begin{equation}
                H_n(I^4 - S^1 \times S^1 \times B^2) \cong 
                    \begin{cases}
                		\mathbb{Z} & n = 0,1,3 \\
                		\mathbb{Z}^2 & n = 2 \\
                		0 & \text{otherwise. }\\
                		\end{cases}
            	\end{equation}
            	Thus, $\beta_n = 1$ for $n=0,1,3$, $\beta_2 = 2$, and all other Betti numbers are~$0$.
    
                The same approach can be used to find $H_n(I^4 - S^1 \times B^3)$. 
                Since the homology groups of the factors of $S^1 \times B^3$ and its boundary $S^1 \times S^2$ are well-known,
                applying the K\"unneth theorem under the same Tor functor conditions that are described above will produce the following results.
                \begin{equation}
                    H_n(S^1 \times B^3) \cong 
                    \begin{cases}
                        \mathbb{Z} & n = 0,1 \\
                        0 & \text{otherwise }\\
                    \end{cases}
                \end{equation}
                \begin{equation}
                    H_n(S^1 \times S^2) \cong 
                    \begin{cases}
                        \mathbb{Z} & n = 0,1,2,3 \\
                        0 & \text{otherwise }\\
                    \end{cases}
                \end{equation}
                The Mayer-Vietoris Sequence can then be set up to determine the homology groups of $I^4 - S^1 \times B^3$, which will yield the following isomorphism.
                \begin{equation}
                    H_n(I^4 - S^1 \times B^3) \cong 
                    \begin{cases}
                        \mathbb{Z} & n = 0,2,3 \\
                        0 & \text{otherwise }\\
                    \end{cases}
                \end{equation}
                Thus, $\beta_n = 1$ for $n=0,2,3$, and all other Betti numbers are $0$. 
                
                Computing the homology groups of $I^4 - S^2 \times B^2$ again follows the same approach and we can reuse the result for $H_n(S^1 \times S^2)$, since $S^1 \times B^3$ and $S^2 \times B^2$ have homeomorphic boundaries. Finally, we can directly apply the Mayer-Vietoris Sequence to compute the homology groups for $I^4 - B^4$ since $H_{n}(B^4)$ and $H_{n}(S^3)$ are well-known.
    
                The results provided in this appendix are summarised in Table~\ref{tab:betti_numbers}.
    
        \section{Supplementary materials}\label{app:supplementary_materials}
            The supplement to this paper comprises several parts. Some of the material is available from the Open Research Newcastle repository at 
            \url{https://doi.org/10.25817/HV0T-V961-7D6C-7B06},
            which contains the 4D dataset that was used to perform the experiments that were outlined in Section~\ref{sec:experiments_and_results_4d} and two 4D visualisations in video format. In addition, the 3D data that was generated for the experiments that were presented in Section~\ref{sec:a_study_using_3d} can be accessed via the supplementary materials link on the ACM Transactions on Graphics webpage for this paper.
        
        \subsection{4D Dataset}\label{app_subsec:dataset}
            The 4D dataset can be downloaded as either a single approximately 30GB .zip file, or as a series of twenty 1.5GB .zip files.
            The dataset parameters can be found in Section~\ref{subsec:dataset_parameters}. Each sample is saved as a NumPy \verb+.npz+ archive file. Each \verb+.npz+ file contains three elements, which are named as follows.
            \begin{enumerate}
                \item `data': the 4D cube,
    
                \item `bettiNumbers': the label containing the Betti numbers, and 
    
                \item `objects': the label containing a count of $B^4$, $S^1 \times B^3$, $S^2 \times B^2$, and $S^1 \times S^1 \times B^2$.
            \end{enumerate}
    
            The archive can be loaded into a Python program by using the NumPy \verb+load+ command, as shown in the following example.\\
            {\small
                \noindent
                \verb+>>> import numpy as np+\\
                \verb+>>> loadedSample = np.load(`4d_mixed_#0_1.npz')+\\
                \verb+>>> cube = loadedSample[`data']+\\
                \verb+>>> labelBetti = loadedSample[`bettiNumbers']+\\
                \verb+>>> labelObjs = loadedSample[`objects']+
            }

        \subsection{Visualising 4D samples}\label{app_subsec:visualising_4d_samples}
    
            By inverting the toxel values of a sample (setting 0 to 1, and vice versa), we are able to visualise the cavities within a sample.
            Two 4D visualisations in video format are included in the supplementary materials, which depict the $128^4$ 4D sample that was presented in Figure~\ref{fig:visualising_a_4d_sample}. The videos present each of the 128 slices along the $w$-axis as a 3D frame, which allows us to see examples of the balls and various tori that have been used to introduce cavities into the interior of a 4D cube.
            The first video depicts each slice from a fixed aspect and the second video uses an aspect that rotates as each slice is presented, which provides a broader view of the features that can be seen within the sample.

            \subsection{3D Dataset}\label{app_subsec:3Ddataset}
            In the 3D data repository, each \verb+.npz+ file contains the following elements.
            \begin{enumerate}
                \item `data': the 3D cube,
    
                \item `bettiNumbers': the label containing the Betti numbers, and 
    
                \item `bettiNumbersInv': the label containing the Betti numbers of the cube's inverse.
            \end{enumerate}

\end{document}